\documentclass[sigconf]{aamas} 

\usepackage{balance} 

\usepackage{amsmath}
\usepackage{mathrsfs} 
\usepackage{bm}

\def\gA{{\mathcal{A}}}

\def\gD{{\mathcal{D}}}

\def\gH{{\mathcal{H}}}
\def\gI{{\mathcal{I}}}

\def\gL{{\mathcal{L}}}

\def\gN{{\mathcal{N}}}

\def\gR{{\mathcal{R}}}
\def\gS{{\mathcal{S}}}
\def\gT{{\mathcal{T}}}
\def\gU{{\mathcal{U}}}

\def\E{{\mathbb{E}}}

\def\valpha{{\bm{\alpha}}}
\def\vmu{{\bm{\mu}}}
\def\vtheta{{\bm{\theta}}}
\def\vpi{{\bm{\pi}}}
\def\vtau{{\bm{\tau}}}
\def\va{{\bm{a}}}

\def\vo{{\bm{o}}}

\def\vr{{\bm{r}}}

\usepackage{subfigure}
\usepackage{xcolor} 
\usepackage[ruled,boxed,linesnumbered]{algorithm2e} 
\usepackage[T1]{fontenc}
\usepackage{aecompl}

\setcopyright{ifaamas}
\acmConference[AAMAS '23]{Proc.\@ of the 22nd International Conference
on Autonomous Agents and Multiagent Systems (AAMAS 2023)}{May 29 -- June 2, 2023}
{London, United Kingdom}{A.~Ricci, W.~Yeoh, N.~Agmon, B.~An (eds.)}
\copyrightyear{2023}
\acmYear{2023}
\acmDOI{}
\acmPrice{}
\acmISBN{}

\acmSubmissionID{63}

\title[AAMAS-2023 Formatting Instructions]{Self-Motivated Multi-Agent Exploration}

\newcommand{\referenceauthors}{Shaowei Zhang, Jiahan Cao, Lei Yuan, Yang Yu, and De-Chuan Zhan}

\author{Shaowei Zhang\textsuperscript{\rm$\star$}}\thanks{$^\star$ Equal contribution.\\
$^\dagger$ Corresponding author}
\affiliation{
  \institution{National Key Laboratory for Novel Software Technology, Nanjing University}
  \city{Nanjing}
  \country{P.R.China}}
\email{zhangsw@lamda.nju.edu.cn}

\author{Jiahan Cao\textsuperscript{\rm$\star$}}
\affiliation{
  \institution{National Key Laboratory for Novel Software Technology, Nanjing University}
  \city{Nanjing}
  \country{P.R.China}}
\email{caojh@lamda.nju.edu.cn}

\author{Lei Yuan}
\affiliation{
  \institution{National Key Laboratory for Novel Software Technology, Nanjing University; Polixir Technologies}
  \city{Nanjing}
  \country{P.R.China}}
\email{yuanl@lamda.nju.edu.cn}

\author{Yang Yu}
\affiliation{
  \institution{National Key Laboratory for Novel Software Technology, Nanjing University; Polixir Technologies}
  \city{Nanjing}
  \country{P.R.China}}
\email{yuy@nju.edu.cn}

\author{De-Chuan Zhan\textsuperscript{\rm$\dagger$}}
\affiliation{
  \institution{National Key Laboratory for Novel Software Technology, Nanjing University; Polixir Technologies}
  \city{Nanjing}
  \country{P.R.China}}
\email{zhandc@nju.edu.cn}

\begin{abstract}
In cooperative multi-agent reinforcement learning (CMARL), it is critical for agents to achieve a balance between self-exploration and team collaboration. 
However, agents can hardly accomplish the team task without coordination and they would be trapped in a local optimum where easy cooperation is accessed without enough individual exploration. Recent works mainly concentrate on agents' coordinated exploration, which brings about the exponentially grown exploration of the state space. 
To address this issue, we propose \emph{Self-Motivated Multi-Agent Exploration} (SMMAE), which aims to achieve success in team tasks by adaptively finding a trade-off between self-exploration and team cooperation. In SMMAE, we train an independent exploration policy for each agent to maximize their own visited state space. Each agent learns an adjustable exploration probability based on the stability of the joint team policy.
The experiments on highly cooperative tasks in StarCraft II micromanagement benchmark~(SMAC) demonstrate that SMMAE can explore task-related states more efficiently, accomplish coordinated behaviours and boost the learning performance.

\end{abstract}


\newcommand{\BibTeX}{\rm B\kern-.05em{\sc i\kern-.025em b}\kern-.08em\TeX}

\begin{document}


\pagestyle{fancy}
\fancyhead{}


\maketitle 


\section{Introduction}
\label{sec:introduction}
Cooperative \emph{multi-agent reinforcement learning} (CMARL) has achieved outstanding results~\cite{oroojlooyjadid2019review,  gronauer2022multi} and has been applied to many real-world applications, such as autonomous driving~\cite{auto_driver}, multi-agent path finding \cite{greshler2021cooperative}, natural language processing tasks~\cite{marl_nlp}, and dynamic algorithm configuration~\cite{madac}. 
Most of the methods can be divided into two categories: the value-based methods~\cite{vdn, qmix, qplex} and others based on policy gradient~\cite{maddpg, coma, mappo}. The value-based methods and their variants~\cite{cao2021linda, DBLP:conf/ijcai/YuanWWZCGZZY22, DBLP:conf/aaai/YuanWZWZ0Z22} can have better performance in complex tasks like the StarCraft II micromanagement benchmark~\cite{smac}. However, they usually neglect efficient exploration, which is a particularly challenging problem in complex scenarios. Agents will be trapped in a local optimum if they only focus on team collaboration and cannot conduct coordinated behaviours if they only pay attention to exploration.

\begin{figure}[t]
    \centering
    \subfigure[3s5z\_vs\_3s6z]{
    \includegraphics[width=0.22\textwidth]{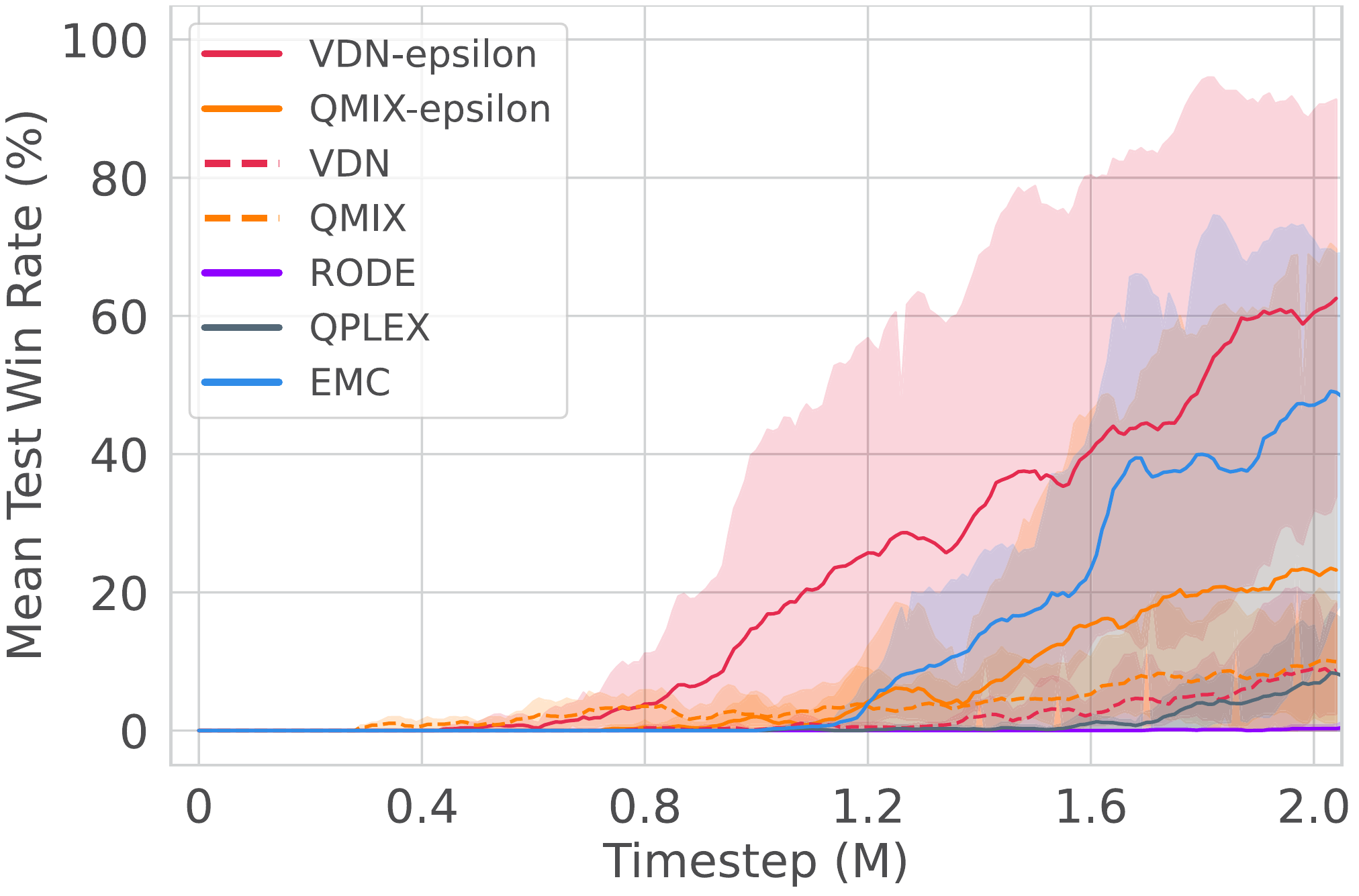}
    }
    \subfigure[6h\_vs\_8z]{
    \includegraphics[width=0.22\textwidth]{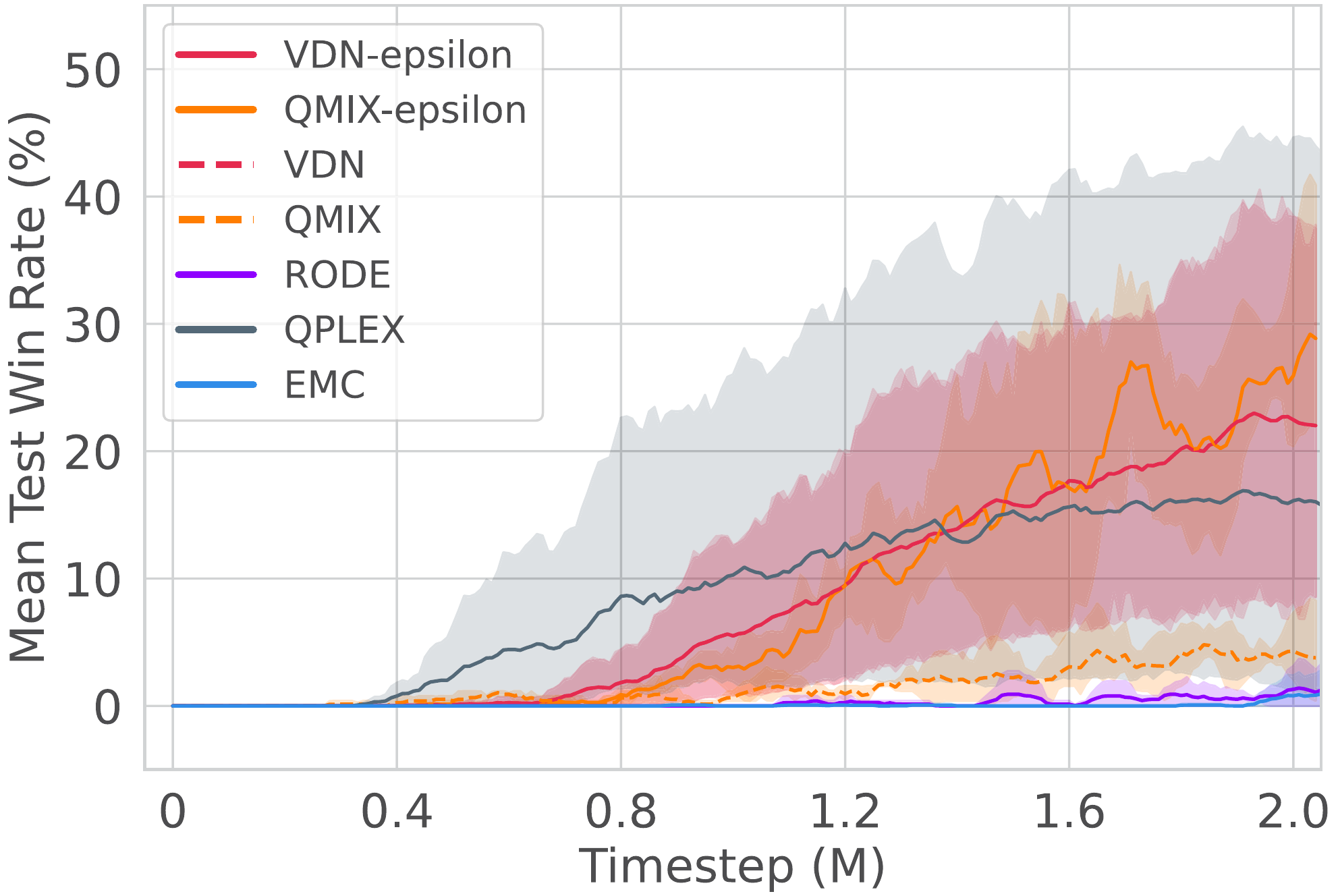}
    }
    \vspace{-0.2in}
    \caption{Results on two super hard maps need strong exploration from SMAC, VDN-epsilon and QMIX-epsilon can promote coordination significantly. }
    \label{fig:vdn_qmix_vs_sota}
    \vspace{-0.3in}
\end{figure}

To address the problem of MARL exploration, a vanilla technique that is commonly used in value-based MARL algorithms is $\varepsilon$-greedy~\cite{gronauer2022multi,exploration_survey1}. Recently, a variety of methods have been proposed and they focus more on coordinated exploration. EITI \& EDTI~\cite{eiti_edti} use the interactions between the agents to measure the collaborative exploration. CMAE~\cite{cmae} focuses on the low-dimensional space that indeed affects the team cooperation reward. EMC~\cite{emc} proposes that the local Q function is affected by the influence of other agents and guides the exploration based on this viewpoint. However, these algorithms ignore individual exploration and consider only joint exploration objective, which has a very large space and is not as efficient as individual exploration.

The mentioned methods can facilitate the MARL exploration ability in someway but are inefficient in complex scenarios, as those methods seldom consider self-exploration from an individual point of view. What's more, they are far away from human coordination. If a person is not familiar with the team behaviour in the current situation, one will first learn how to cooperate with others. Otherwise, one will explore more unknown situations to cooperate better with others in more situations~\cite{klein1993decision}. Although those mentioned MARL exploration approaches can achieve 
coordination improvements in some MARL tasks by designing complex modules, we find that we can achieve competitive results with vanilla MARL methods such as QMIX~\cite{qmix}, by simply focusing on self-exploration. 

For example, in the widely-used benchmark SMAC~\cite{smac} almost all existing MARL methods fail to succeed in 2 million steps in the challenging scenarios such as maps 3s5z\_vs\_3s6z and 6h\_vs\_8z, where sufficient exploration is required~\cite{vdn, qmix, qplex, rode, emc}. However, in our experiments, by adjusting the degree of self-exploration, we find that QMIX~\cite{qmix} and VDN~\cite{vdn} can achieve significant improvements in exploration and can succeed in super hard tasks~(Figure~\ref{fig:vdn_qmix_vs_sota}).
Concretely, here the self-exploration is adjusted by changing the $\varepsilon$ end value of $\varepsilon$-greedy~(Figure~\ref{fig:epsilon_finish_value_show} in Appendix~\ref{sec:epsilon_finish_ablation}) which starts with the value 1.0 and decays linearly with time to a fixed end value $\varepsilon_T$ in the popular implementation of QMIX~\cite{smac}. The results show that the $\varepsilon$ value has a strong association with the degree of exploration, and the probability that the agent adopts an exploration policy at one step during training time, should be applied at different timesteps during training to trade off between self-exploration and team cooperation to for high coordination.  However, to achieve this, it needs extensive work for hyper-parameter tuning at each timestep, and it is not tractable to manually decide which $\varepsilon$ value is suitable at each timestep in complex tasks.  

Towards designing a method that can adaptively adjust the degree of exploration at different timesteps, we propose a novel multi-agent exploration method \emph{Self-Motivated Multi-Agent Exploration} (SMMAE). Inspired by the widely used uncertainty measurement from single agent Reinforcement Learning~\cite{smm, mega_omega}, we posit that learning a proper uncertainty about the multi-agent system and employing it to promote individual exploration can facilitate coordination in CMARL. 
In a multi-agent system,  when the uncertainty between agents' actions and others' observations is limited, the agents should explore individually to jump out of the local optimum. On the contrary, the agents' lack of awareness of others makes it hard to achieve coordinated behaviors when the uncertainty in the multi-agent system is high, and the agents should explore less and learn how to cooperate first before exploring more. Specifically, the uncertainty of the joint policy for the agents can be measured using the correlation between each agent's action and observations of other agents. We then use mutual information~\cite{kraskov2004estimating} to denote the correlation. The observation of other agents is used to predict the current agent's action, and the cross entropy loss is used as a criterion. The smaller the cross entropy loss is, the less the uncertainty is.
Our main contributions are:
\vspace{-0.2in}
\begin{itemize}
    \item We study the exploration probability in MARL and reveal that the suitable $\varepsilon$ value at different timestep has a huge valuable influence on exploration behaviours and final performance, especially in complex environments.
    \item SMMAE can adaptively adjust exploration probability according to the uncertainty in the multi-agent system to trade off between self-exploration and team cooperation.
    \item The experiments demonstrate the strong ability of SMMAE to explore task-related state space efficiently.
\end{itemize}

\section{Background}
\vspace{-0.05in}
\subsection{Cooperative Multi-Agent System Model}
This paper considers a fully cooperative multi-agent system, where all the agents need to cooperative with each other to earn a shared team reward. The system can be modelled by a Dec-POMDP~\cite{decpomdp} tuple $G = \langle \gN, \gS, \gA, P, \gR, \Omega, O, n, \gamma \rangle$, where $\gN$ is a finite set of $n$ agents, $s \in \gS$ is a global state of the environment in the set of possible states, $\gA$ is the finite action set and $\gamma \in [0, 1)$ is the discount factor. Due to the partially observable settings, agent $i \in \gN$ is only accessible to a local observation $o_i \in \Omega$ according to the observation function $O(s, i)$. Each agent has an observation-action trajectory history $\tau_i \in \gT \equiv \left( \Omega \times \gA \right)^*$. At each timestep the joint action of the team is $\va = \langle a_1, \cdots, a_n\rangle \in \boldsymbol{\gA} \equiv \gA^n$ where $a_i \in \pi_i(a \mid \tau_i)$ is an action selected by each agent $i$. The team joint action will lead to the next global state $s'$ according to the environment transition function $P(s'\mid s, \va)$ and the team will earn the shared reward $r = \gR(s, \va)$. The joint policy $\vpi \equiv \langle \pi_1, \cdots, \pi_n \rangle$ aims to maximize the joint value function $V^\vpi(s) = \E_{s_0 : \infty} \left[ \sum_{t=0}^\infty \gamma^t r_t \mid s_0 = s, \vpi\right]$, which induces a joint action-value function $Q^\vpi(s, \va) = \gR(s, \va) + \gamma \E_{s'} \left[ V^\vpi (s') \right]$.

\vspace{-0.1in}
\subsection{Centralized Training with Decentralized Execution (CTDE) paradigm}
Centralized Training with Decentralized Execution (CTDE)~\cite{ctde_1} has been a popular paradigm for cooperative multi-agent reinforcement learning. In a CTDE paradigm, during training the global states are available for all the agents by using a centralized controller, while during test time each agent has to use its local network to select an action based on its local trajectories. Deep Q Network~\cite{dqn_nature} and its derivatives~\cite{double_dqn, dueling_dqn} have achieved great performance in reinforcement learning tasks, especially in game tasks. In the multi-agent system, the tuple in replay buffer $\gD$ is $\left(\vtau_t, \va_t, r_t, \vtau_{t+1}\right)$ and the joint action-value function is $Q_{tot}(\vtau_t, \va_t; \vtheta)$, where  $\vtheta$ are the parameters of the Q network and will be learnt by the following Temporal Difference~(TD) error~\cite{sutton2018reinforcement}:
\begin{equation}
    \begin{split}
    &\gL\left( \vtheta \right) \\
    &=
    \E_{(\vtau_t,\va_t, r_t, \vtau_{t+1}) \sim \gD} \left[ r + \gamma \max_{\va_{t+1}} Q_{tot}(\vtau_{t+1}, \va_{t+1}; \vtheta^-) - Q_{tot}(\vtau_t, \va_t; \vtheta) \right]^2,
    \end{split}
\end{equation}
where $\vtheta^-$ are the parameters of the target network and will be updated by $\vtheta$ periodically. Considering the CTDE paradigm, many works~\cite{vdn, qmix, wqmix, qplex} adopt the decomposition structure between the joint action-value function $Q_{tot}$ and local action-value function $Q_i$, and obtain good performance.

\vspace{-0.15in}
\section{Related Work}
\vspace{-0.05in}
\subsection{ Exploration in Reinforcement Learning} 
Various methods for exploration have been studied in single-agent reinforcement learning~\cite{exploration_survey1}. Some works explore by focusing on the environment dynamics. ICM~\cite{icm} adopts both forward and inverse models to build a good feature space for curiosity and explore what is controlled by or affects the agent. Pathak et al.~\cite{exploration_via_disagreement} propose to use an ensemble of dynamics models. They use the variance over the output of these networks in the ensemble to induce the exploration.
There are some works focused on the environment novelty to induce exploration. Some works are the generation of count-based methods and use density models which can generate pseudo-counts of visited states to measure the uncertainty of the agents~\cite{unifying_cb_exploration, ostrovski2017count}. Some works use prediction error to reflect the novelty of the states. RND~\cite{rnd} uses a fixed randomly initialized network to get the embedding of the state, then uses a learnable network to reconstruct this embedding, and the reconstruction loss will be adopted as an intrinsic reward to guide exploration. Based on this, NovelD~\cite{noveld} proposes to use the difference of RND novelty between adjacent time step as the intrinsic reward, and adds a restriction in an episode so that the agent could earn a reward only when it visits the states for the first time. Therefore, NovelD prefers to explore unexplored boundary states, thereby exploring more valuable state regions. SMM~\cite{smm} aims to learn a state marginal distribution that matches the prior distribution and uses this as the target of exploration.
Some works adopt epsilon schedule for better exploration. Tokic~\cite{adaptiveepsilongreedy} uses the difference between value functions to adjust the size of epsilon. When the agent's knowledge becomes certain about the environment, the degree of exploration will be reduced. $\varepsilon z$-greedy~\cite{epsilonzgreedy} replaces a single action with a sequence of actions (option), choosing an option instead of choosing a random action with probability $\varepsilon$.
Some works focus on noise for better exploration. NoisyNet~\cite{noisynet} replaces the fully connected value network with a learnable noise network. Plappert et al.~\cite{parameter_space_noise} add adaptive-scale noise to the parameters of the policy. 
There are also some works that study exploration from some interesting perspectives. Eysenbach et al.~\cite{diversity_is_all_you_need} use diversity-driven exploration to learn distinguishable and diverse skills,
while SMiRL~\cite{smirl} reduces the entropy of the visited states during exploration to exclude negative effects of perturbations and acquire complex behaviours and skills without supervision.
C-BET~\cite{c_bet} first learns a exploration policy across environments without extrinsic rewards, then transfers the learned exploration policy to the target tasks. P{\^\i}slar et al.~\cite{when_should_agents_explore} study the problem when the agent should explore and reveal the results of the methods with different types of exploring temporal structure. Agent57~\cite{agent57} adopts a more engineering approach. The Q network is divided into an intrinsic part and an extrinsic part. NGU is a policy that treats different degrees of exploration equally~\cite{ngu}. Based on NGU, Agent57 uses a meta-controller to select the policy adaptively. 

\vspace{-0.15in}
\subsection{Multi-Agent Exploration}
There are also many studies on exploration in multi-agent reinforcement learning. To achieve committed exploration, MAVEN~\cite{maven} adopts hierarchical control and the policies of the agents are conditioned on the shared latent variable generated by a hierarchical policy. Wang et al.~\cite{eiti_edti} propose two exploration methods, EITI and EDTI, to induce cooperative exploration by capturing the influence of one agent's on other agents. EITI quantifies the influences on the state transition dynamics, while EDTI quantifies both transition and reward influences. However, they are not scalable as they need to use a approximation way to measure the influence of other agents on one agent when there are many agents, which can cause the method to fail. CMAE~\cite{cmae} proposes that reward function only depends on a small subset of the large state space. Therefore, it first explores in the projected low-dimensional space of the high-dimensional state space, then they select goals from the low-dimensional space and train the exploration policies to reach the goal to explore higher-dimensional space continuously. However, it's hard to find the effective low-dimensional projection in complex MARL tasks. EMC~\cite{emc} proposes that local Q function of each agent can capture the novelty of states and the influences between agents. To induce coordinated exploration, EMC proposes to use prediction errors of individual Q function as the intrinsic reward. However, EMC is also not scalable, as it has to maintain a huge episodic memory buffer.

\vspace{-0.1in}
\section{Methodology}
\begin{figure*}[t]
  \centering
  \vspace{-0.1in}
  \includegraphics[width=0.87\linewidth]{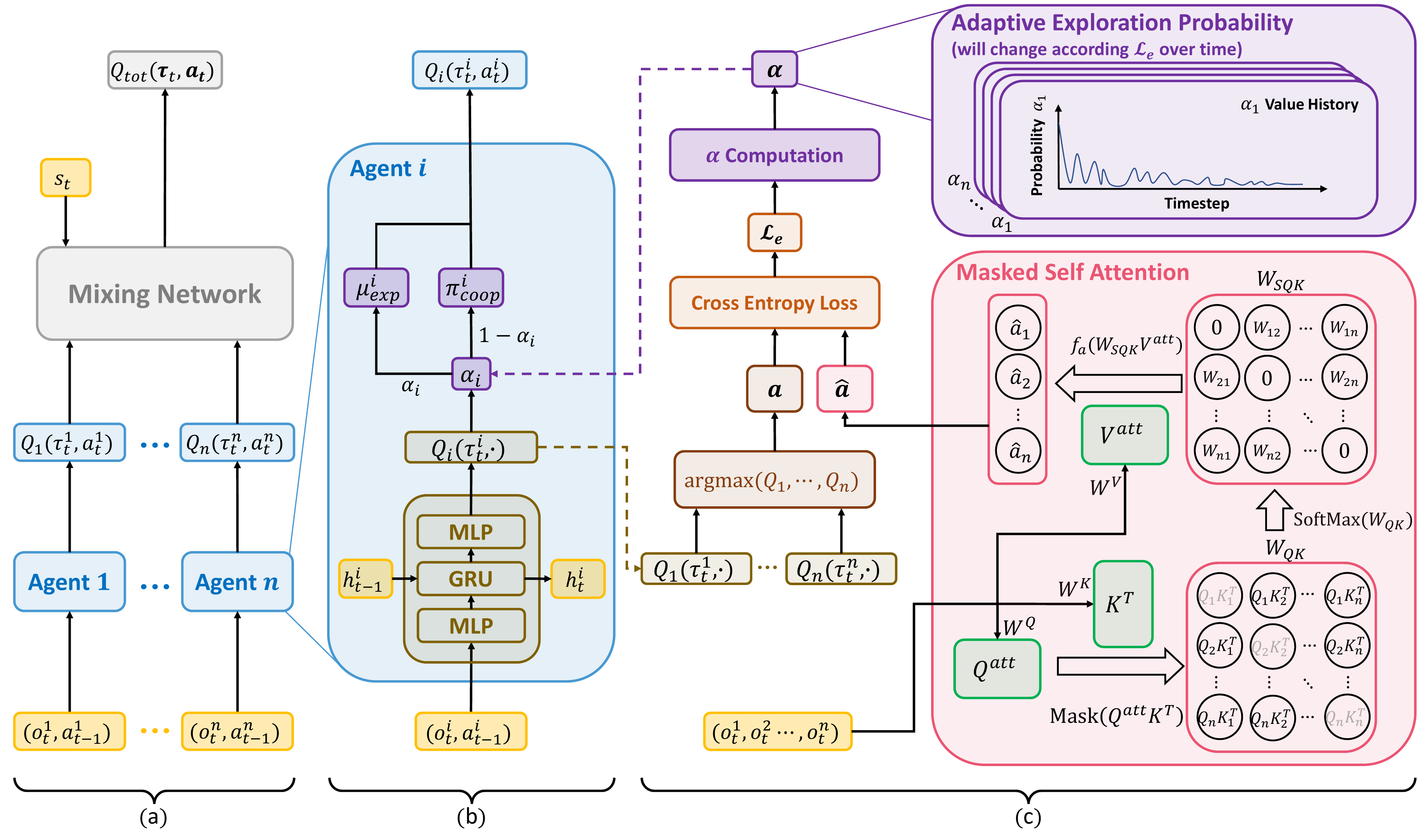}
  \vspace{-0.15in}
  \caption{SMMAE framework. (a) Overall structure. (b) Agent network structure using $\alpha_i$ as the exploration probability and $\mu_{exp}^i$ as the exploration policy. (c) Network structure to compute adaptive exploration probability $\valpha$.} 
  \vspace{-0.2in}
  \label{fig:framework}
\end{figure*}

In this section, we introduce \emph{Self-Motivated Multi-Agent Exploration}~(SMMAE), a novel method for effective exploration in MARL. Figure~\ref{fig:framework} shows the whole SMMAE framework. Part (a) and (b) in Figure~\ref{fig:framework} are the common structures in Centralized Training with Decentralized Execution~(CTDE) algorithms using value decomposition. Each agent $i$ uses its observation $o_t^i$ and action $a_{t-1}^i$ as input and outputs local Q-value to select action. The mixing network takes all the outputs of the agents to train the multi-agent policy during training time. SMMAE mainly changes the structure of each individual agent in part~(b), and the main contribution of our method is illustrated in part~(c).

\vspace{-0.1in}
\subsection{Adaptive Exploration}
In $\varepsilon$-greedy, the value of $\varepsilon$ will decay linearly to a fixed end value $\varepsilon_T$ in a common implementation. However, the experiments prove that exploration probability with fixed end value is not efficient (Figure~\ref{fig:vdn_qmix_vs_sota}).
Therefore, we adopt adaptive exploration probability. As the exploration probability $\varepsilon$ is for all agents, we use $\valpha = \langle \alpha_1, \cdots, \alpha_n  \rangle$ to denote the exploration probabilities of $n$ agents, where $\alpha_i$ represents the probability of agent $i$ to select the exploration policy. 

We associate $\valpha$ with the uncertainty of the multi-agent system~(Part (c) in Figure~\ref{fig:framework}). The agents cooperate well when the uncertainty of the multi-agent system is limited and they should increase the exploration intensity to jump out of the local optimum by increasing $\alpha_i$. When the system uncertainty is high, the agents lack awareness of others and are hard to achieve coordinated behaviours. Under this condition, agents should take more exploitation to reduce the uncertainty by decreasing $\alpha_i$.

The uncertainty of the multi-agent system can be measured by the correlation between agents. When the correlation between agent $i$'s action $a_i$ and the other agents' observation $\vo_{-i}$ is low, the uncertainty of the multi-agent system is high because the agents focus more on their own action selection than the team cooperation. We propose to use the mutual information $\gI (a_i, \vo_{-i})$ to reflect the correlation between the agent $i$'s action $a_i$ and the other agents' observation $\vo_{-i}$:

\vspace{-0.2in}
\begin{equation}
    F(\vo, \va)
        =\sum_{i=1}^{n} \gI\left(a_{i}, \vo_{-i}\right)
        =\sum_{i=1}^{n} \gH\left(a_{i}\right)-\gH\left(a_{i} \mid \vo_{-i}\right),
\end{equation}
\vspace{-0.1in}

where $\gH$ denotes entropy. Then we can derive a lower bound for the team mutual information term using a variational estimator:

\vspace{-0.1in}
\begin{equation}
    \label{eq:team_correlation}
    \begin{split}
        &F(\vo, \va) \\
        &=\sum_{i=1}^{n} \E_{a_i \sim p\left( a_i \mid \vo_{-i} \right)}\left[\log p\left(a_{i} \mid \vo_{-i}\right)\right]- \E_{a_i \sim p\left(a_i \right)} \left[\log p\left( a_{i} \right) \right]  \\
        &\geq \sum_{i=1}^{n} \E_{a_i \sim p\left( a_i \mid \vo_{-i} \right)} \left[\log q_{\xi_{i}}\left(a_{i} \mid \vo_{-i}\right)\right] - \E_{a_i \sim p\left(a_i \right)} \left[\log p\left(a_{i}\right)\right],
    \end{split}
\end{equation}
\vspace{-0.15in}

where $q_{\xi_{i}}\left(a_{i} \mid \vo_{-i}\right)$ is a variational posterior estimator of $p\left(a_{i} \mid \vo_{-i}\right)$ with parameter $\xi_i$. As $p(a_i)$ is the prior of actions, the last term is a constant and can be ignored. Therefore, we only need to consider the first term.
We use a network to represent $q_{\xi_{i}}\left(a_{i} \mid \vo_{-i}\right)$, where the input is $\vo_{-i}$ and the output is $\hat{a}_i$. Then the first term  $ \E_{a_i \sim p\left( a_i \mid \vo_{-i} \right)} \left[\log q_{\xi_{i}}\left(a_{i} \mid \vo_{-i}\right)\right]$ is the network's negative cross entropy loss $\gL_{ce}(a_i, \hat{a}_i \mid \vo_{-i})$.

Based on this, we use a two-level heuristic method to adjust $\alpha_i$:
\begin{equation}
    \label{eq:update_alpha}
    \begin{split}
        \alpha_i^{new} =
        \begin{cases}
            \alpha_{low}~~~,~~~~~~~~~~\gL_{ce}(a_i, \hat{a}_i \mid \vo_{-i}) \geq \gL_{threshold}\\
            \min \left( \alpha_i^{old} +\dfrac{\alpha _{high}-\alpha _{low}}{\lambda_{\alpha}},\alpha_{high} \right) ,~~else,
        \end{cases}
    \end{split}
\end{equation}
where $\alpha_{low}$ and  $\alpha_{high}$ are the lower bound value and upper bound value, respectively. $\lambda_{\alpha}$ is the scaling parameter for $\valpha$ increasing steps and it takes $\lambda_{\alpha}$ steps to increase from the lower bound to the upper bound, and $\gL_{threshould}$ is the loss threshold. For the policy convergence, $\alpha_{high}$ will decrease linearly to $\alpha_{low}$ at the end of the training. For stability, we update $\valpha$ every $E_\alpha$ episodes.

We need to mask agent $i$'s observation as we use other observation $\vo_{-i}$ of agent $i$ to predict agent $i$'s action. It means computing $\gL_{ce}(a_i, \hat{a}_i \mid \vo_{-i})$ for each agent $i$ requires optimizing $n$ networks simultaneously and will consume large amounts of computing resources. Instead, we utilize the structure of the attention mechanism~\cite{transformer} to estimate all the $n$ losses within one network:
\begin{equation}
    g^{att}(\vo_t)= \texttt{SoftMax}\left( \frac{\texttt{Mask}\left(Q_{t}^{att} (K_{t}^{att})^T \right)}{\sqrt{d_k}}\right) V_{t}^{att},
\end{equation}
where $Q_{t}^{att}= \vo_t W^Q$, $ K_{t}^{att}=\vo_t W^K$, $V_{t}^{att}=\vo_t W^V$ and $W^Q, W^K, W^V$ denote the parameter matrix for the attention mechanism. Here $d_k$ is the attention key size. As we use other observation $\vo_{-i}$ of agent $i$ to predict, the function $\texttt{Mask}(X)$ will mask the correlation matrix and set the diagonal of the matrix to 0 (Figure~\ref{fig:framework}).  And the action prediction is: 
$\hat{\va}_t = f_a(g^{att}(\vo_t))$, where $f_a$ is an action prediction network with three fully-connected layers.

\vspace{-0.1in}
\subsection{Explore by Maximizing State Entropy}
Existing algorithms only focus on the global exploration policy, which will face the curse of dimensionality. Instead, we propose to use individual exploration and each agent can maximize its own exploration space, which is more efficient. SMMAE learns $n$ independent exploration policies $\vmu_{exp} \equiv \langle \mu_{exp}^1, \cdots, \mu_{exp}^n \rangle$ for agents. Inspired by SMM~\cite{smm}, we use state marginal matching to maximize individual exploration space. In MARL, each agent only has access to local observation, so we use a local observation to approximate a local state and match the visited observation distribution $\rho_{\pi_i}(o^i)$ with a target distribution $p^*(o^i)$, where

\vspace{-0.1in}
\begin{equation}
    \begin{split}
        &\rho_{\pi_i}(o^i) \\
        &=\E_{a_t^i \sim \pi_i(a_t^i \mid o_t^i), o^i_{t+1}\sim O(P(s_{t+1}\mid s_t,\va_t))}\left[ \frac{1}{T}\sum_{t=1}^T \bm{1}(o^i_t=o^i)\right],
    \end{split}
\end{equation}
\vspace{-0.1in}

and $p^*(o^i)$ is obtained using prior information. As there is not any prior information in our experiments, the target distribution of the exploration policy is uniform. To match the two distributions, for each policy we aim to optimize the following objective:
\begin{equation}
    \begin{split}
         &\min_{\mu_{exp}^i} D_{\texttt{KL}}\left(\rho_{\mu_{exp}^i}(o^i) \| p^*(o^i) \right)\\
    &= \max_{\mu^i_{exp}} \E_{o^i \sim \rho_{\mu^i_{exp}}} \left[ \log p^*(o^i) \right] + \gH_{\mu^i_{exp}}(o^i).
    \end{split}
\end{equation}
 As $p^*(o^i)$ is uniform, to optimize the objective, each exploration policy only needs to maximize the last term $\gH_{\mu^i_{exp}}(o^i)$ and uses the following reward
\begin{equation}
    \vr_{exp}(\vo_t, \va_t) \triangleq \lambda_{exp} \vr_{env} -\log q_{\xi_{exp}}\left(\vo_{t+1}\right)
    \label{eq:exp_reward}
\end{equation}
to learn the exploration policy, where $\vr_{env} \equiv \langle r_{env}, \cdots, r_{env}\rangle$ and $\lambda_{exp}$ is the scaling parameter for environment reward. We use environment reward here to assist the exploration. Here $\vr_{exp}(\vo_t, \va_t) \equiv \langle r^1_{exp}(o^1_t, a^1_t), \cdots, r^n_{exp}(o^n_t, a^n_t)\rangle$,  and the reward for the exploration policy of agent $i$ is
\begin{equation}
    r^i_{exp}(o^i_t, a^i_t) \triangleq \lambda_{exp} r_{env} -\log q^i_{\xi_{exp}}\left(o^i_{t+1}\right),
\end{equation}
where $q^i_{\xi_{exp}}(o_{t+1}^i)$ is the variational estimator for the visiting probability of agent $i$'s observation $o^i$ w.r.t $\rho_{\mu_{exp}^i}(o^i)$. In practice, each $q^i_{\xi_{exp}}(o_{t+1}^i)$ is estimated by a Variational Auto-Encoder~(VAE)~\cite{vae} model, where the input and the reconstruction objective are both $o^i$. The reconstruction loss of the VAE model is used as the last term $-\log q^i_{\xi_{exp}}\left(o^i_{t+1}\right)$.

\begin{algorithm}[t]
    \caption{SMMAE}
    \SetKwInput{init}{Init}
    \init{exploration probability $\valpha$, exploration policy $\vmu_{exp}$, multi-agent Q-learning cooperation policy $\vpi_{coop}$, exploration replay buffer $\gD_{exp}$, cooperation replay buffer $\gD_{coop}$}
    \BlankLine
    \For{episode $= 1, \cdots, E$}{
        Update $\valpha$ every $E_\valpha$ episodes according to Eq.~\ref{eq:update_alpha} \\
        Reset the environment. Get state $s_{1}$ and observations $\vo_{1}=\left(o_{1}^{1}, \cdots, o_{1}^{n}\right)$ \\
        \For{$t=1, \cdots, T$}{
            \For{$i=1, \cdots, n$}{
                \eIf{$x \sim \gU(0,1)<\alpha_i$}{
                    Use exploration policy to select action $a_{t}^{i} \sim \mu_{exp}^{i}\left(\tau_{t}^{i}\right)$ \\
                }{
                    Use cooperation policy to select action $a_{t}^{i} \sim \pi_{coop}^{i}\left(\tau_{t}^{i}\right)$ \\
                }
            }
            Set joint action $\boldsymbol{a}_{t}=\left(a_{t}^{1}, \cdots, a_{t}^{n}\right)$ \\
            The environment takes a step. Get $r_{env}, s_{t+1}, \vo_{t+1}$\\
            Compute $\vr_{e x p}$ according to Eq.~\ref{eq:exp_reward}\\
            Add transition tuples $\left\{\left(s_{t}, o_{t}^{i}, a_{t}^{i}, s_{t+1}, o_{t+1}^{i}, r_{exp}^{i}\right) \mid i=1, \cdots, n\right\}$ to $\mathcal{D}_{\text {exp }}$\\
            Add transition tuple $\left\{\left(s_{t}, \boldsymbol{o}_{t}, \boldsymbol{a}_{t}, s_{t+1}, \boldsymbol{o}_{t+1}, r_{env}\right) \right\}$ to $\mathcal{D}_{\text {coop }}$\\
        }
        Update density models $q_{\xi_{exp}}^1, \cdots, q_{\xi_{exp}}^n$ using $\left(\vo_1, \cdots, \vo_T \right)$\\
        Train exploration policy $\vmu_{exp}$ using $\mathcal{D}_{e x p}$ \\
        Train cooperation policy $\vpi_{coop}$ using $\mathcal{D}_{\text {coop }}$
    }
    \label{alg:main_alg}
\end{algorithm}

We use the buffer $\gD_{coop}$ for multi-agent Q-learning and an extra buffer $\gD_{exp}$ for exploration policy learning to record the tuple $\left\{\left(s_{t}, o_{t}^{i}, a_{t}^{i}, s_{t+1}, o_{t+1}^{i}, r_{exp}^{i}\right) \mid i=1, \cdots, n\right\}$, where the trajectories in $\gD_{exp}$ are the same as that in $\gD_{coop}$. The TD loss for all exploration policies $\vmu_{exp}$ is:
\begin{equation}
    \begin{split}
        \gL_{exp} =& \frac{1}{n}\sum_{i=1}^n \gL^i_{exp}(\vtheta^i_{exp})\\
        =& \frac{1}{n}\sum_{i=1}^n\E_{({\tau}^i_t,{a}^i_t,r^i_t,{\tau}^i_{t+1}) \sim \gD_{exp}}\left[\left( r_t^i \right.\right.\\
        +& \left.\left. \gamma_{exp} \max_{a} Q_{exp}({\tau}^i_{t+1}, {a} ; \vtheta_{exp}^{i-}) -  Q_{exp}({\tau}^i_{t}, {a}^i_{t} ; \vtheta^i_{exp}) \right)^2\right],
    \end{split}
\end{equation}
where $\mu^i_{exp}$ is the exploration policy of agent $i$, $\vtheta^{i-}_{exp}$ are the parameters of the target network and will be updated by $ \vtheta^i_{exp}$ periodically. All the exploration policy $\vmu_{exp} $ will be trained after the multi-agent Q-learning policy $\vpi_{coop}$ is trained.

\begin{figure*}[ht]
    \centering
	\centerline{\includegraphics[width=0.95\linewidth]{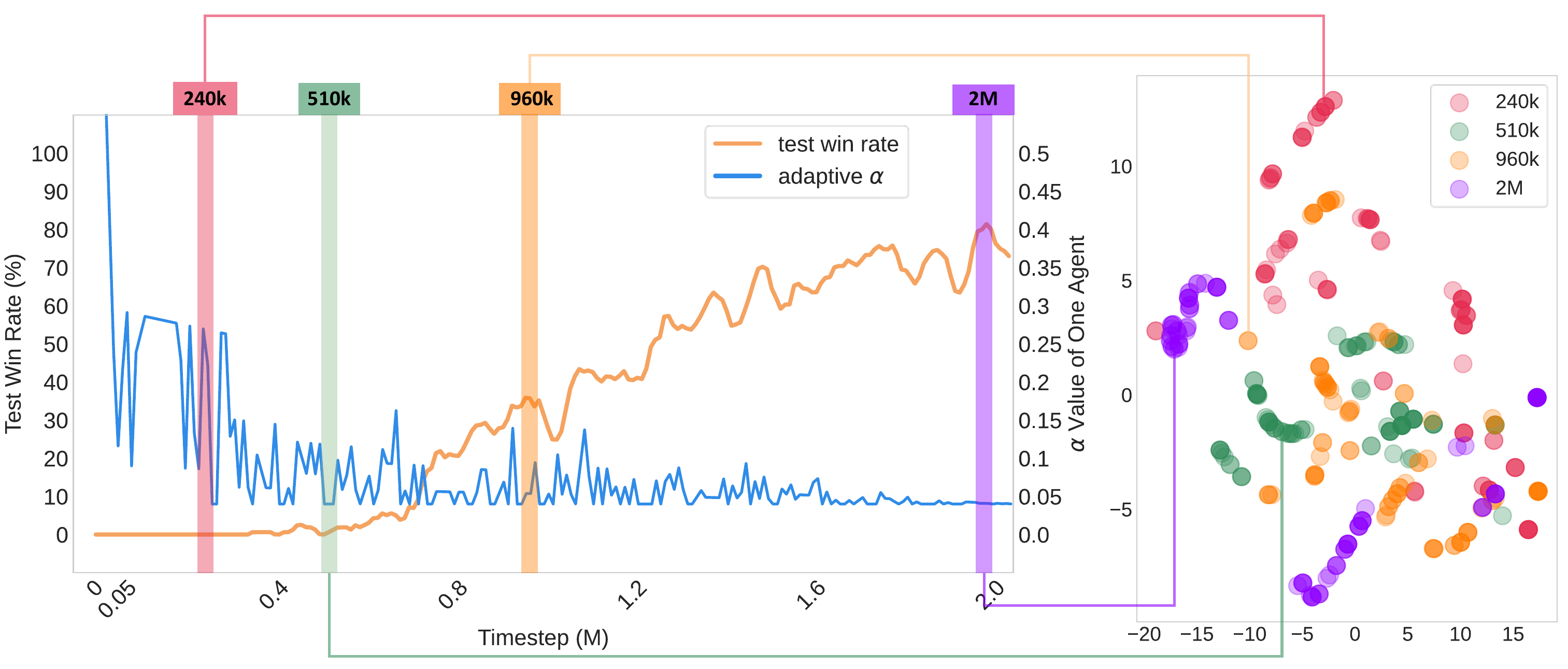}}
	\vspace{-0.1in}
	\caption{Influence of adaptive exploration probability on \emph{6h\_vs\_8z}. The \emph{left} part shows the correlation between the performance curve of self team and the adaptive $\alpha$ curve of one agent during 2 million training steps.
	The \emph{right} part visualizes the agent's observations (t-SNE projection on 2-D) during the training process, which illustrates the dynamic exploration preference and its impact on the performance.}
	\label{fig:adaptive_alpha_show}
\end{figure*}

The whole algorithm is summarized by Algorithm~1. 
We induce the adaptive exploration probability based on the uncertainty of the multi-agent system, and then adopt a new individual exploration policy. We first describe the adaptive way to control the exploration probability, which uses the correlation between action and observation of the agents as the criterion~(lines~2 and 4-11). Then we introduce the individual exploration policy, which is optimized by state marginal matching~\cite{smm}~(lines 14-15 and 18-19).

\vspace{-0.1in}
\section{Experiments}
\label{sec:experiments}

We conduct experiments in this section to validate the efficiency of SMMAE, and we benchmark it on QMIX~\cite{qmix} for it's widely proved coordination ability in MARL\footnote{The experiments are based on \href{https://github.com/oxwhirl/pymarl}{PyMARL} framework. Code can be found at https://github.com/Zhang-Shaowei/SMMAE.}. Implementation details of SMMAE are shown in Appendix~\ref{sec:imple_detail}.
We first select a training process of SMMAE and study the effects of adaptive exploration probability on the observation areas of the agents during this training process. Then, we choose the benchmark SMAC~\cite{smac} as the testbed to study the efficient exploration of SMMAE in complex MARL tasks.\footnote{The SMAC version in our experiments is SC2.4.6.2.69232.} We compare SMMAE with several methods, including EMC~\cite{emc} that focuses on multi-agent exploration and some other baselines~\cite{vdn,qmix,rode,qplex,maven}. After that, we conduct ablation experiments to elaborate the effectiveness of each module. We also show the efficient exploration ability of SMMAE in a visual way. Finally, we show SMMAE is a general framework for MARL algorithms, and it can be applied for other environments like Level Based Foraging~(LBF)~\cite{lbf}. More experimental results can be found in the appendix part.
\begin{figure*}[t]
\centering
    \begin{minipage}[c]{0.72\linewidth}
    \vspace{-0.1in}
	\centerline{\includegraphics[width=0.96\linewidth]{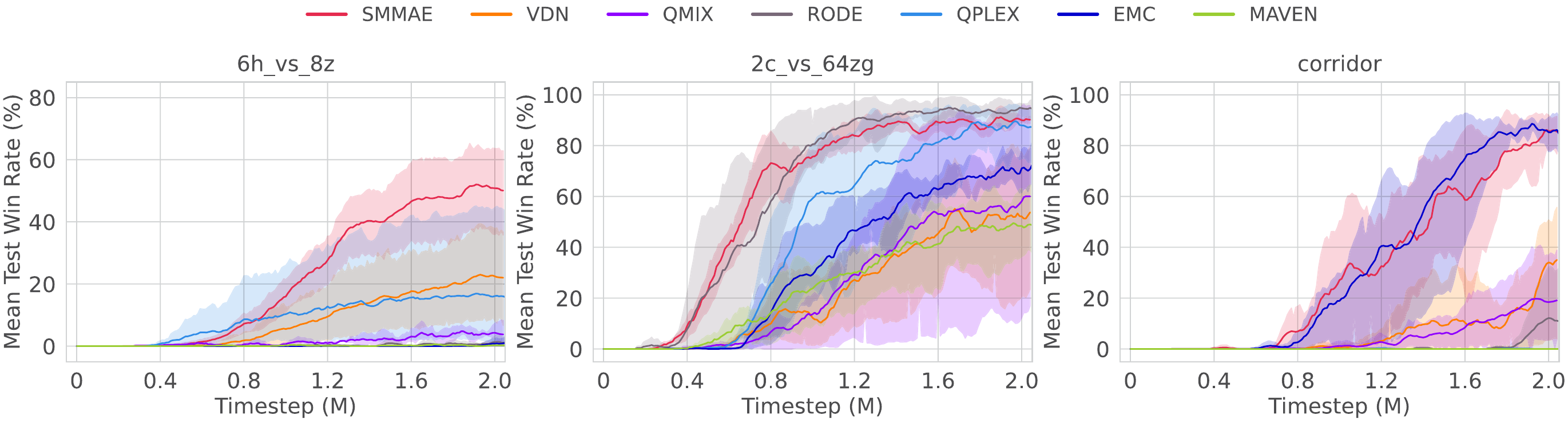}}
	\caption{Results on three super hard maps in the SMAC.}
	\label{fig:smac_sota1}
	\vspace{-0.1in}
	\end{minipage}
	\hfill
    \begin{minipage}[c]{0.26\linewidth}
        \vspace{0.1in}
        \includegraphics[width=0.92\textwidth]{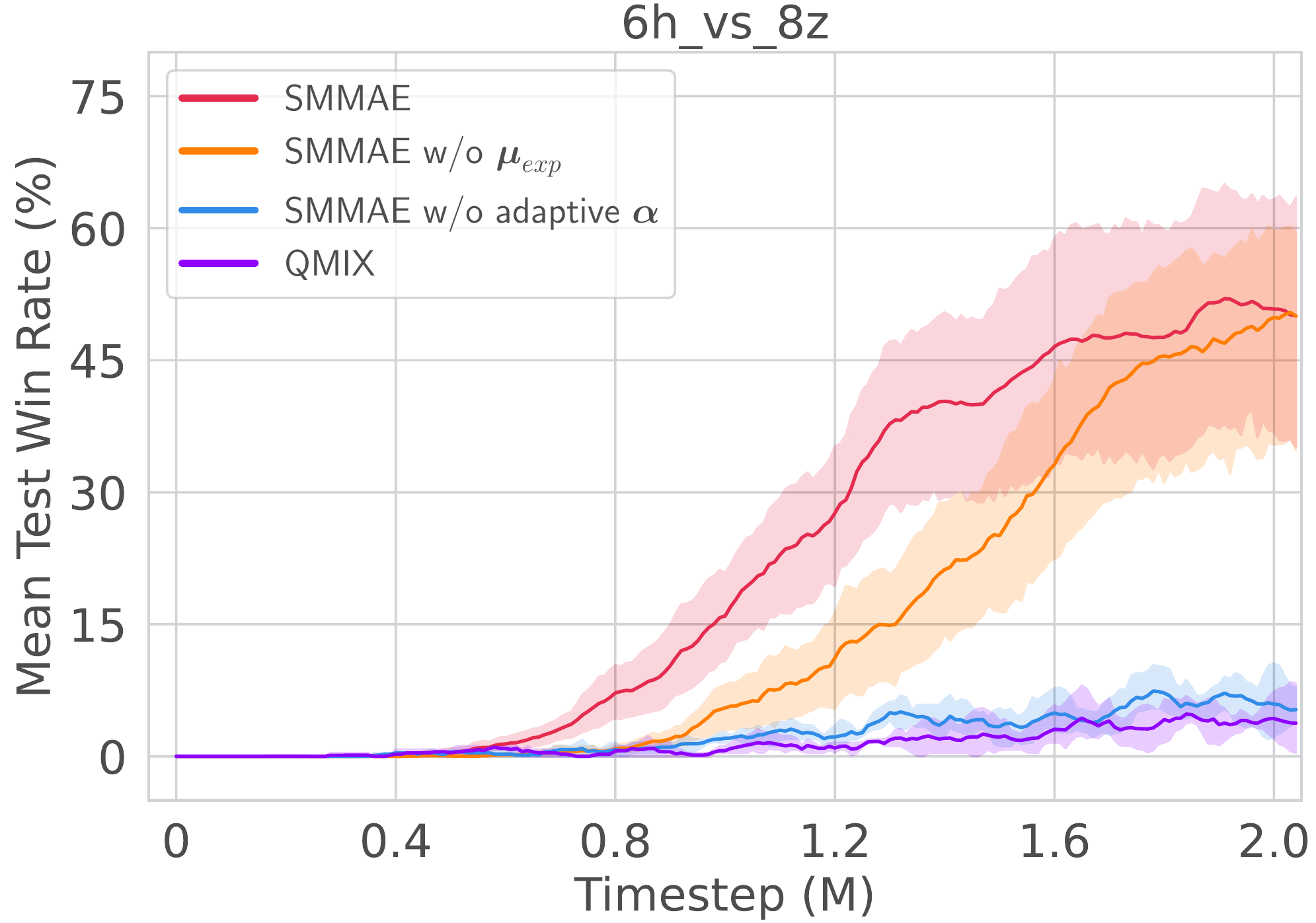}
        \caption{Ablation study.}
        \label{fig:performance_ablation_on_6h_vs_8z}
    \end{minipage}
    \vspace{-0.1in}
\end{figure*}

\vspace{-0.1in}
\subsection{Study on Adaptive Exploration}
In this section, we analyze the effectiveness of adaptive exploration probability. To study the impact of adaptive exploration probability on one agent and the whole team, we experiment on the super hard map \emph{6h\_vs\_8z} in SMAC, and a training process is shown in Figure~\ref{fig:adaptive_alpha_show}. 

The \emph{left} figure of Figure~\ref{fig:adaptive_alpha_show} shows the adaptive exploration probability of one agent and the team test win rate changing over time. We find that the exploration probability of this agent shows a fluctuating decrease (the \emph{blue} curve) as the training proceeds, which is adaptive compared with the vanilla  $\varepsilon$-greedy.
At the beginning of training, the agent explores with a high probability most of the time, while in the middle of training, the exploration probability of the agent varies continuously between high and low values. At the end of the training, the exploration probability tends to converge and gradually stabilizes at a low value.

To analyse the effect of exploration probability on observation areas, we select 4 timesteps, $240k$, $510k$, $960k$, and $2M$, where the exploration probability at $240k$ and $510k$ timesteps are the two peaks, while the exploration probability at $510k$ and $2M$ timesteps are the two valleys. All observations of this agent in the 10 episodes near these 4 timesteps are sampled (four shadow areas in the \emph{left} figure of Figure~\ref{fig:adaptive_alpha_show}), and 100 points were randomly selected from them. The \emph{right} figure in Figure~\ref{fig:adaptive_alpha_show} shows the visualization of these points after we apply t-SNE~\cite{tsne} for dimensionality reduction. We find that for the two timesteps with higher exploration probability, there are more scattered clusters~($240$ and $960k$, \emph{red} and \emph{orange} for each), while for the two timesteps with lower exploration probability, there are fewer clusters ($510k$ and $2M$, \emph{green} and \emph{violet}).

In addition, we can find some correlation between the \emph{brown} test win rate curve and the \emph{blue} exploration probability curve in the \emph{left} figure of Figure~\ref{fig:adaptive_alpha_show}. The \emph{red} shadow area in the figure is in the early stages of training, where the test win rate remains $0\%$ and the agent focuses on exploring and experiencing more state areas. The \emph{green} shadow area in the figure has a low probability of exploration. At these timesteps, the correlation between the agents is low, so the agents are more focused on learning how to cooperate with other agents than exploring. Then, some improvement in the \emph{brown} test win curve can be noticed in this region. The \emph{orange} shadow area in the figure is a peak of exploration. At this time, the correlation between the agents is already high, so the agent learns to explore more areas, thus jumping out of the local optimum. Therefore, there is a significant drop in the test win rate near this area. In the \emph{violet} region, the probability of exploration is close to convergence and the agent focuses on better cooperation. The test win rate reaches the peak of the entire training process, validating the effectiveness of our design of adaptive exploring probability.

\vspace{-0.1in}
\subsection{Performance on SMAC Super Hard Maps}

In order to test the effectiveness of SMMAE, we use the benchmark SMAC~\cite{smac} as the testbed. We compare SMMAE with some baselines, where VDN~\cite{vdn} and QMIX~\cite{qmix} are popular baselines, RODE~\cite{rode} and QPLEX~\cite{qplex} are the state-of-the-art baselines, and MAVEN ~\cite{maven} and EMC~\cite{emc} are the latest methods on MARL exploration. For evaluation, we carry out each experiment with 5 random seeds, and the results are shown with a $95\%$ confidence interval.

\begin{figure}[t]
    \centering
    \begin{minipage}[c]{\linewidth}
        \begin{minipage}[c]{0.49\linewidth}
            \centering 
            \subfigure[c][t-SNE projection. ]{
             \includegraphics[width=0.95\textwidth]{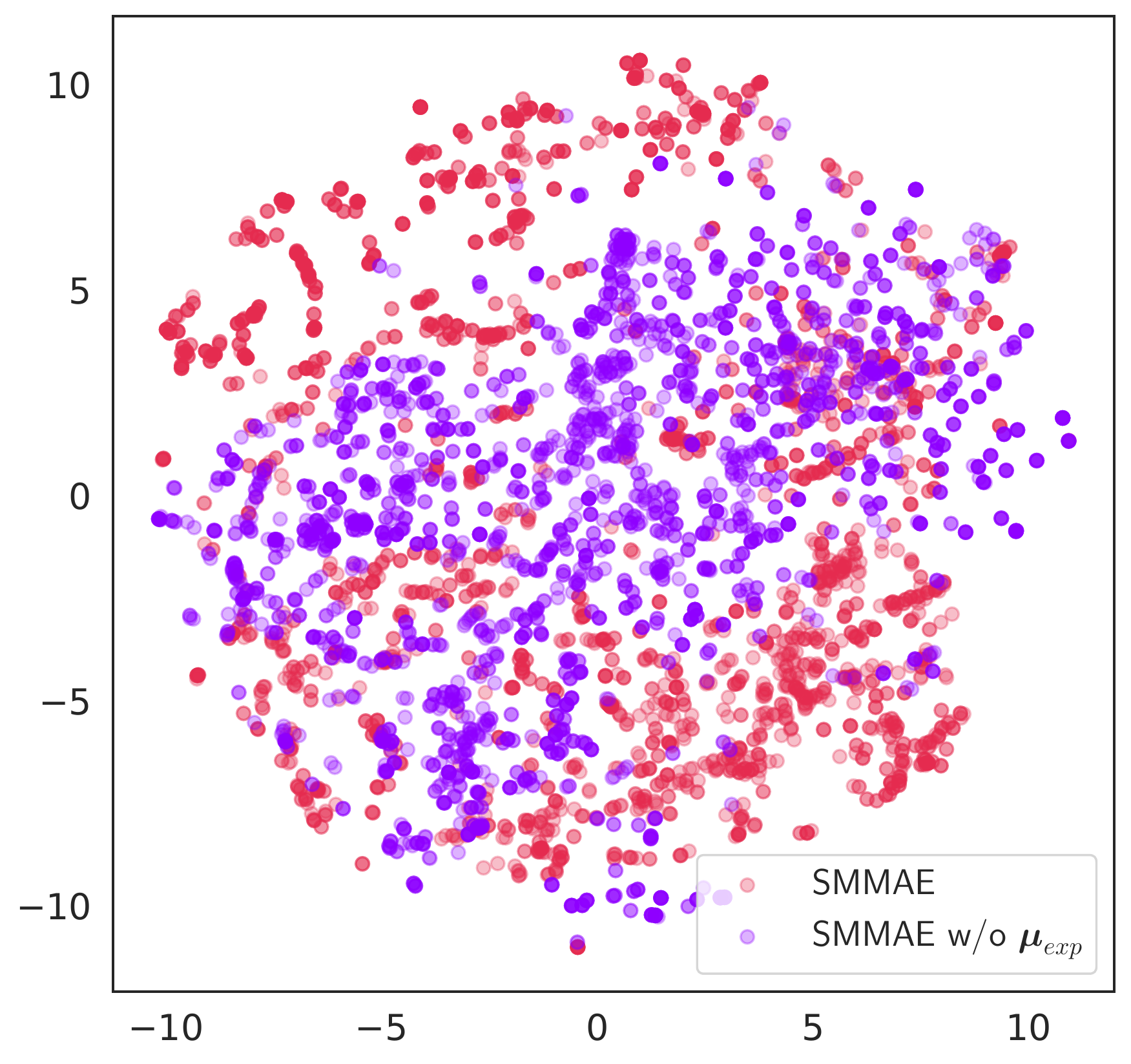}
            \label{fig:two_dimension_tsne_ablation_SMMAE_vs_no_exppolicy}}
        \end{minipage} 
        \begin{minipage}[c]{0.49\linewidth}
            \subfigure[c][Density of one-dimensional.]{
                \includegraphics[width=0.99\textwidth]{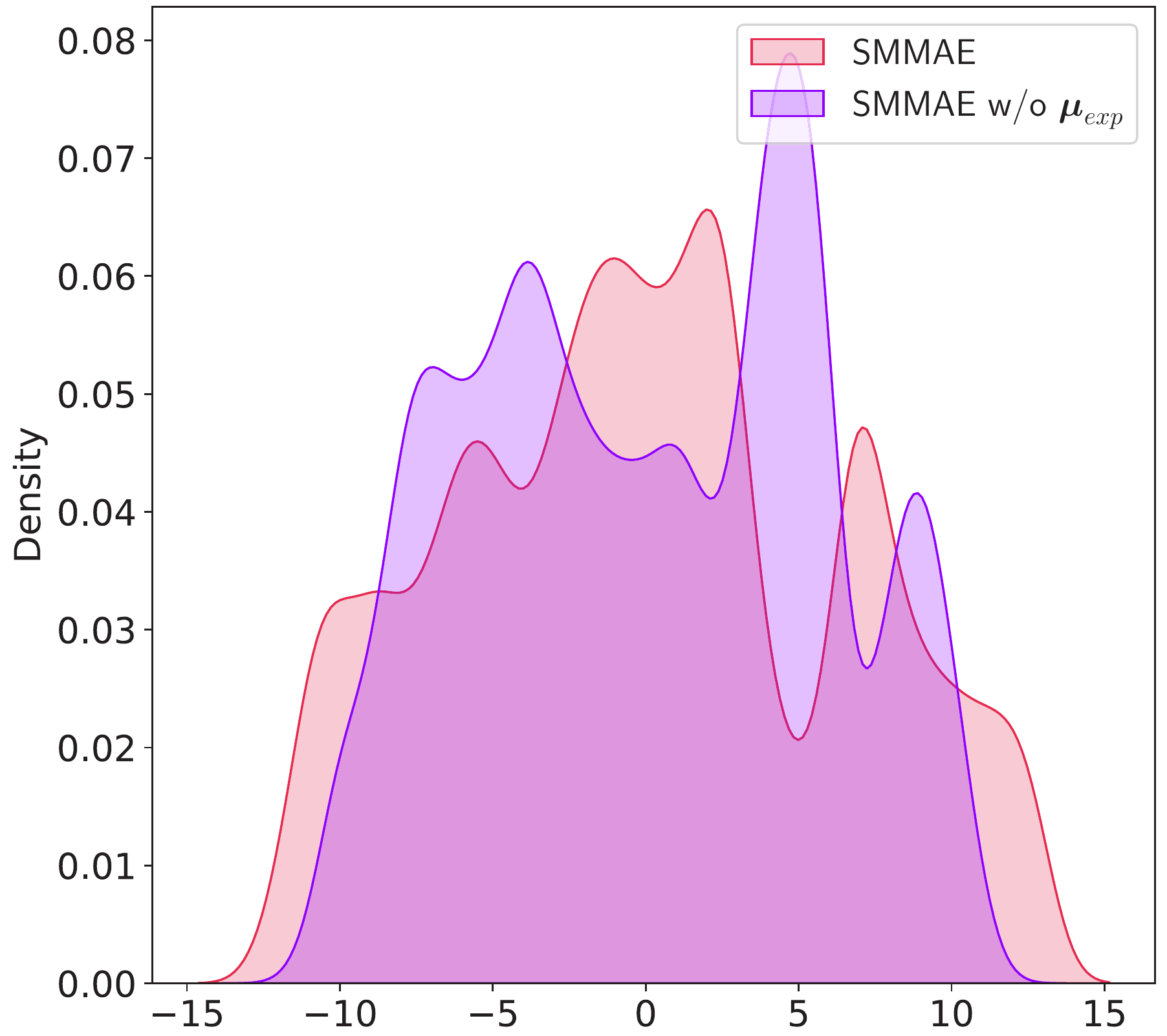}
                \label{fig:one_dimension_tsne_ablation_SMMAE_vs_no_exppolicy}}
        \end{minipage}
        \vspace{-0.1in}
        \caption{The t-SNE projection on global state.}
        \label{fig:tsne_ablation}
        \vspace{-0.2in}
    \end{minipage}
\end{figure}

Figure~\ref{fig:smac_sota1} shows the training curves of SMMAE and other methods on several super hard tasks of SMAC. It can be seen that SMMAE achieves competitive results on tasks such as \emph{6h\_vs\_8z} and \emph{corridor} that require sufficient individual exploration and team cooperation. It shows that adaptive exploration can also achieve the results of other methods using complex modules. This also illustrates the effectiveness of SMMAE because it performs individual exploration and task-related exploration~(Section~\ref{sec:task_related_space} gives a visual explanation).

The final test win rate of SMMAE and that of baseline EMC on \emph{corridor} are almost the same, but it should be noted that EMC needs to use 15GB of GPU when training \emph{corridor}, while our SMMAE only needs about 4GB.

\vspace{-0.05in}
\subsection{Ablation Study}
\label{sec:ablation_study}

As our method includes multiple modules, we compare SMMAE with three ablated methods for ablation study:
\begin{itemize}
    \item SMMAE w/o $\vmu_{exp}$: SMMAE using random exploration instead of the exploration policy $\vmu_{exp}$. 
    \item SMMAE w/o adaptive $\valpha$: SMMAE using fixed ending $\valpha$ instead of the adaptive $\valpha$.
    \item QMIX: the QMIX baseline~\cite{qmix}.
\end{itemize}

\begin{figure}[t]
    \centering
    \begin{minipage}{\linewidth}
        \begin{minipage}[c]{0.48\linewidth}
            \centering 
            \subfigure[c][2c\_vs\_64zg]{
             \includegraphics[width=0.95\textwidth]{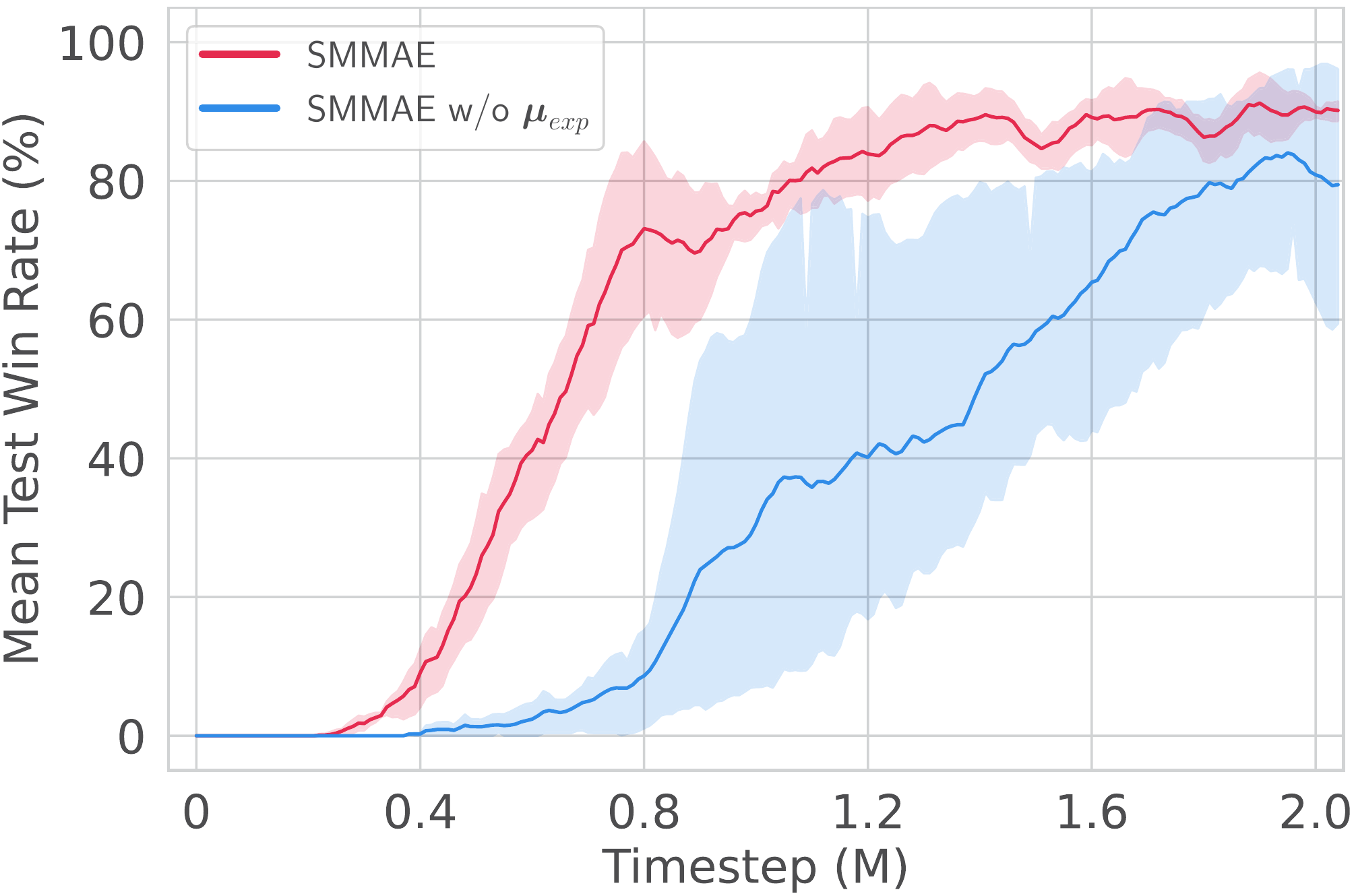}
            \label{fig:performance_ablation_on_2c_vs_64zg}}
        \end{minipage} 
        \begin{minipage}[c]{0.48\linewidth}
            \centering 
            \subfigure[c][corridor]{
             \includegraphics[width=0.95\textwidth]{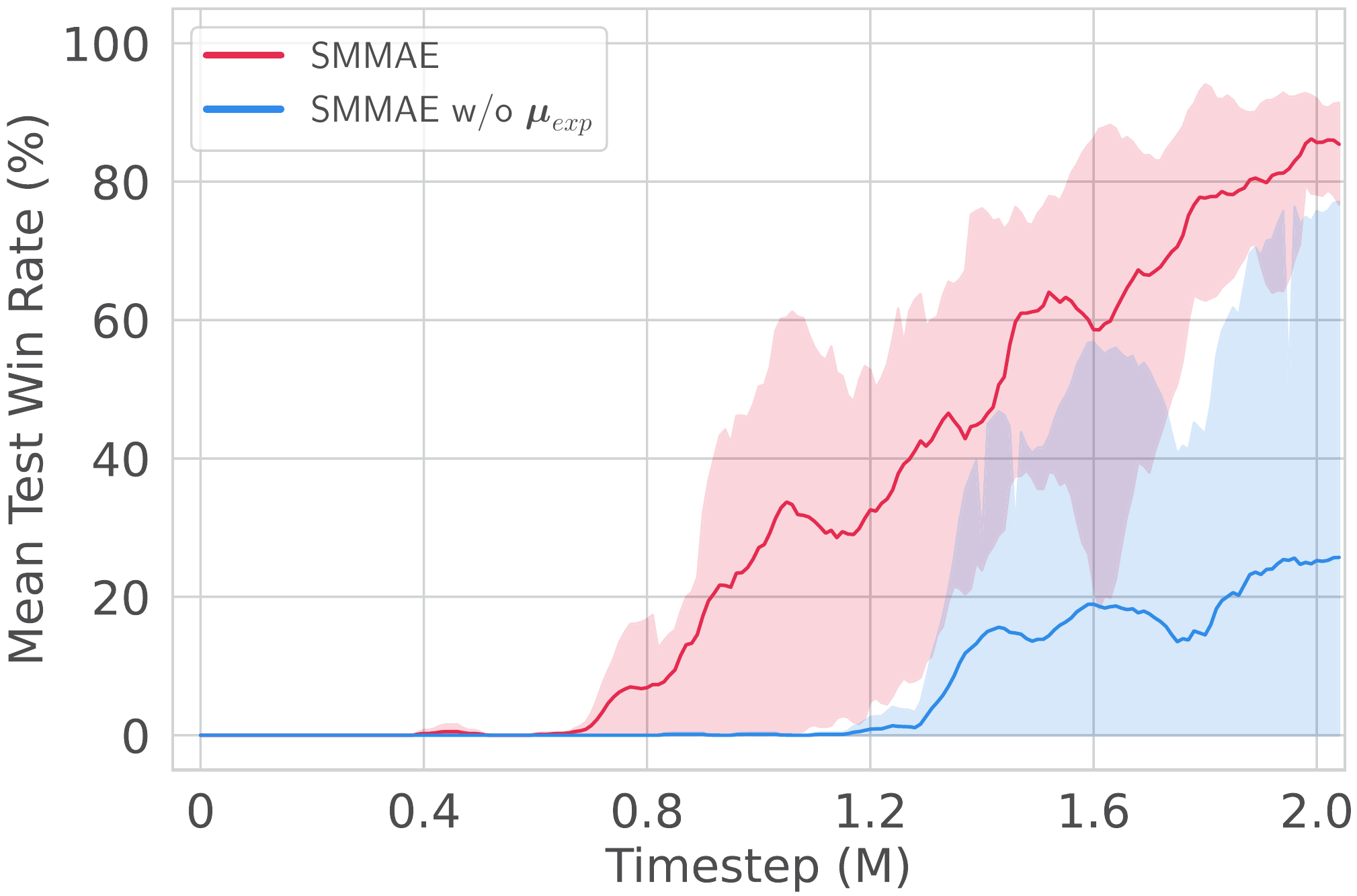}
            \label{fig:performance_ablation_on_corridor}}
        \end{minipage}
        \vspace{-0.15in}
        \caption{More exploration policy ablation in SMAC.}
        \label{fig:appendix_performance_ablation_on_smac}
    \end{minipage}
    \vspace{-0.2in}
\end{figure}

Figure~\ref{fig:performance_ablation_on_6h_vs_8z} shows the average test win rate of these four approaches on the \emph{6h\_vs\_8z} scenario of SMAC. Compared with the original QMIX, SMMAE w/o adaptive $\alpha$ is slightly better, while SMMAE w/o $\vmu_{exp}$ has greatly improves performance. It illustrates the effectiveness of the two modules. Among these four methods, SMMAE achieves the best results, which indicates the effectiveness of using two modules simultaneously. However, it is also observed that the variance of SMMAE becomes larger compared to SMMAE w/o $\vmu_{exp}$, which indicates that increasing the exploration capability of the individual exploration policy may introduce some instability to the training process. Compared with SMMAE w/o $\vmu_{exp}$, SMMAE has similar convergence win rate. Therefore, We conducted experiments on two additional maps \emph{2c\_vs\_64zg} and \emph{corridor}  in SMAC to further investigate the effect of the exploration policies $\vmu_{exp}$ in SMMAE~(Figure~\ref{fig:appendix_performance_ablation_on_smac}). All of these results demonstrate that although random exploration can sometimes perform well, $\vmu_{exp}$ can explore more efficiently and speed up training .

\begin{figure*}[t]
    \centering
        \includegraphics[width=\linewidth]{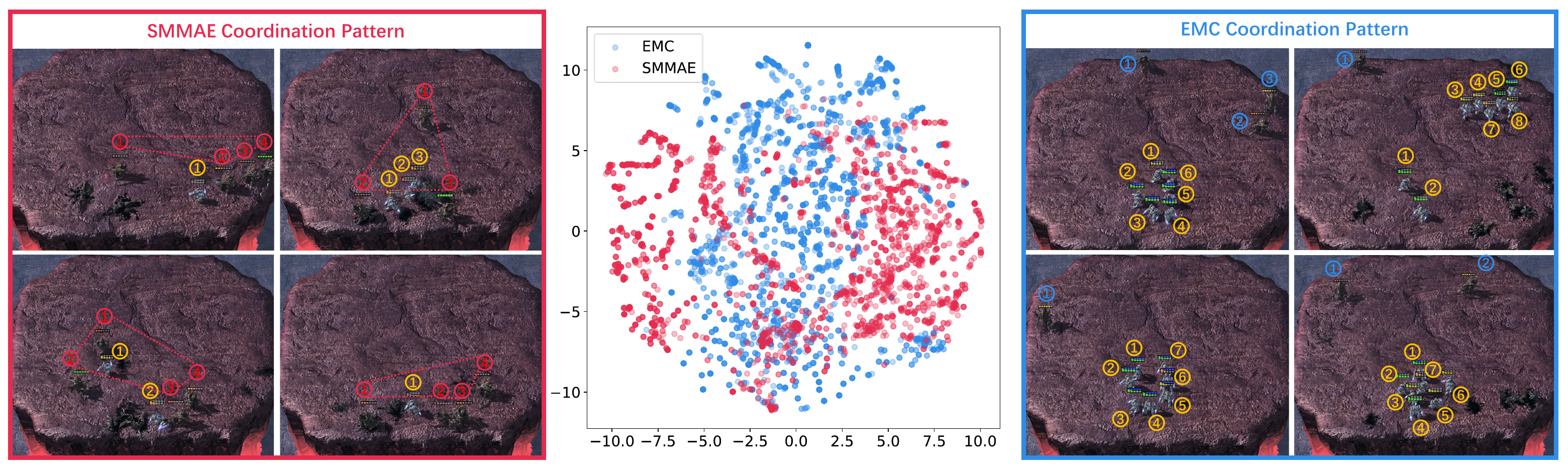}
        \vspace{-0.2in}
    	\caption{Exploration experiments on \emph{6h\_vs\_8z}. \emph{Left}: SMMAE replay during test, where the red ones are controlled by SMMAE and the orange ones are the enemies. \emph{Middle}: Two-dimensional global state t-SNE embeddings during training. \emph{Right}: EMC replay during test, where the blue ones are controlled by EMC and the orange ones are the enemies.}
    	\label{fig:tsne_SMMAE_vs_EMC}
    	\vspace{-0.1in}
\end{figure*}

To qualitatively analyze the exploration ability of the exploration policy, in the ablation experiment, we sample the episodes experienced by SMMAE and SMMAE w/o $\vmu_{exp}$. Then we sample global state points randomly from them, and visualize the points after dimensionality reduction using t-SNE~\cite{tsne}. Figure~\ref{fig:tsne_ablation} shows the results. It can be found that SMMAE has more dispersed clusters in the two-dimensional t-SNE embedding space~(Figure~\ref{fig:two_dimension_tsne_ablation_SMMAE_vs_no_exppolicy}), and SMMAE has more uniform density and wider range in one-dimensional t-SNE embedding space~(Figure~\ref{fig:one_dimension_tsne_ablation_SMMAE_vs_no_exppolicy}). It shows that SMMAE, which uses the exploration policy $\vmu_{exp}$ that maximizes the state entropy of local visited observations, has a stronger exploration ability. Then we try to express the visualization results quantitatively. Because the state space is continuous, we retain two decimal places for each dimension of each global state and calculate an approximate entropy of the visited state based on state counting. The entropy of SMMAE is $15.238$, and the entropy of SMMAE w/o $\vmu_{exp}$ is $14.982$. The quantitative results are consistent with the visualization results~(Figure~\ref{fig:tsne_ablation}).

\begin{figure}[ht]
    \centering
    \begin{minipage}{\linewidth}
        \begin{minipage}[c]{0.48\linewidth}
            \centering 
            \subfigure[c][3s5z\_vs\_3s6z]{
         \includegraphics[width=\textwidth]{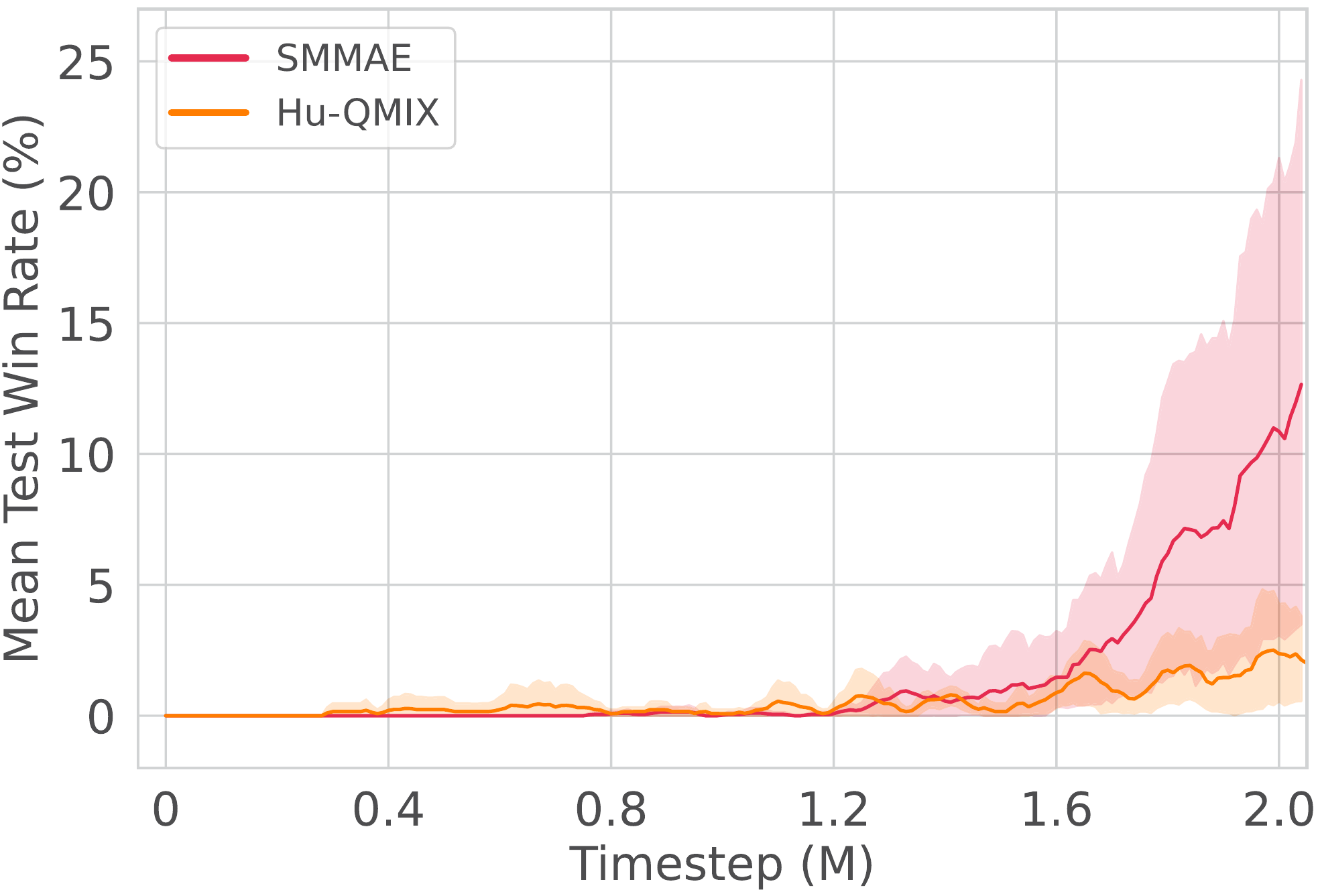}
            \label{fig:smac_vs_hujian_on_3s5z_vs_3s6z}}
        \end{minipage} 
        \begin{minipage}[c]{0.48\linewidth}
            \centering 
            \subfigure[c][6h\_vs\_8z]{
         \includegraphics[width=\textwidth]{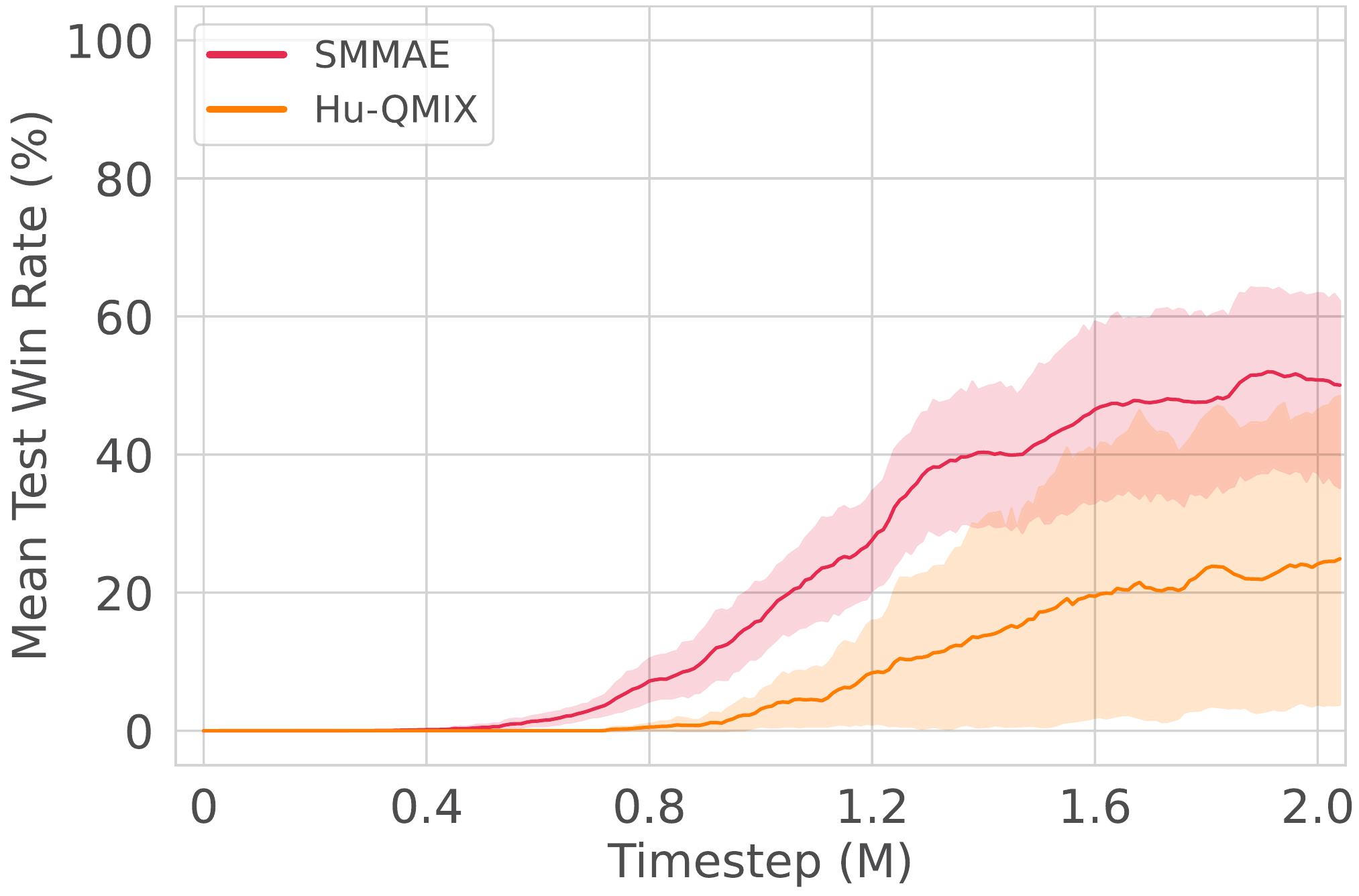}
            \label{fig:smac_vs_hujian_on_6h_vs_8z}}
        \end{minipage} 
        \begin{minipage}[c]{0.48\linewidth}
            \centering 
            \subfigure[c][corridor]{
         \includegraphics[width=\textwidth]{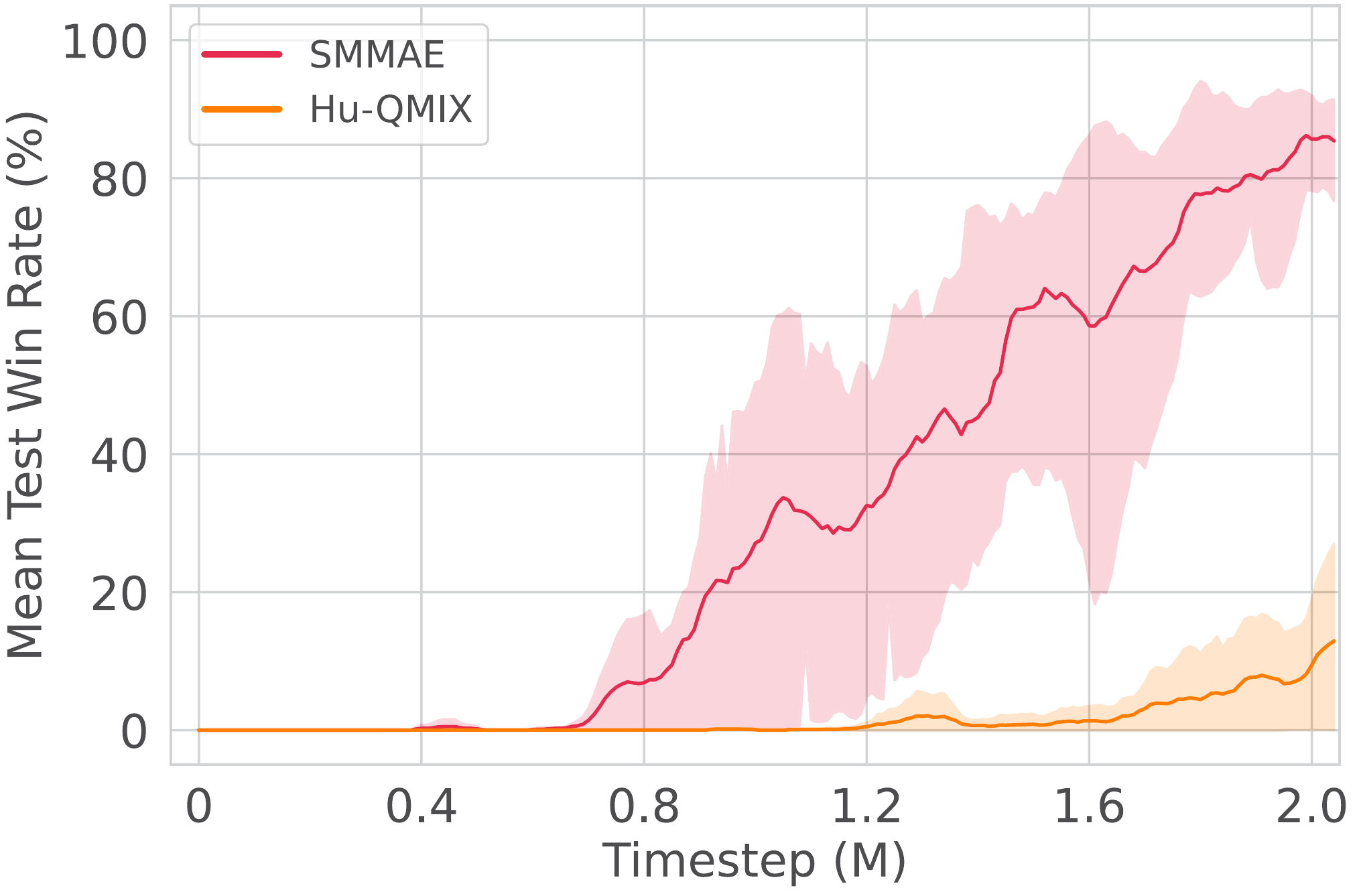}
            \label{fig:smac_vs_hujian_on_corridor}}
        \end{minipage} 
        \begin{minipage}[c]{0.48\linewidth}
            \centering 
            \subfigure[c][MMM2]{
         \includegraphics[width=\textwidth]{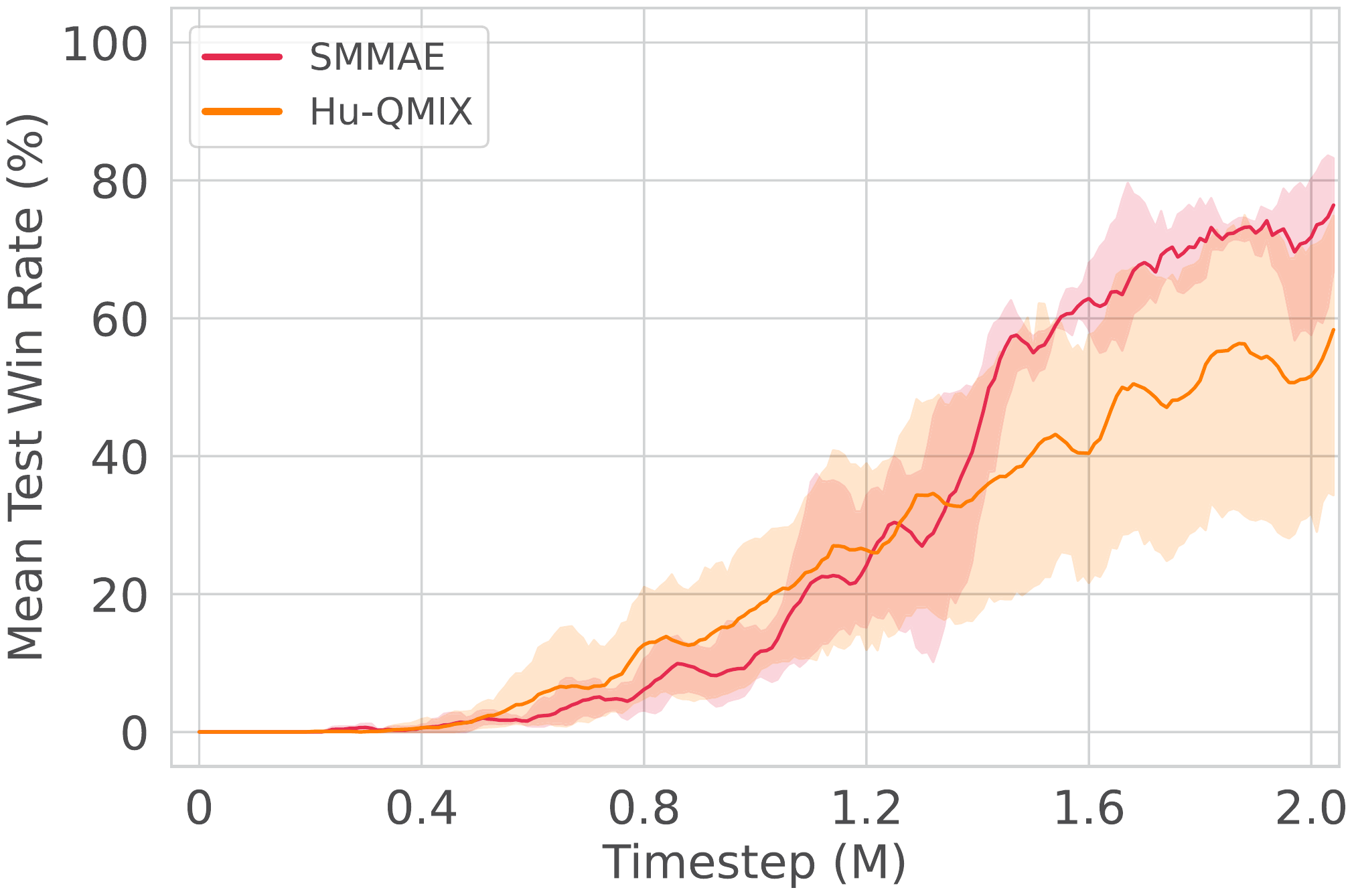}
            \label{fig:smac_vs_hujian_on_MMM2}}
        \end{minipage} 
        \vspace{-0.1in}
        \caption{Comparation with the finetuned QMIX on four super hard maps of SMAC.}
        \label{fig:appendix_smac_vs_hujian}
    \end{minipage}
    \vspace{-0.1in}
\end{figure}

\subsection{Comparation with Finetuned-QMIX}
Hu et al.~\cite{rethinkingqmix} finetune QMIX~\cite{qmix} by adding a variety of implementation tricks and achieve good performance on SMAC. To show the necessity of the designs in SMMAE, we compare SMMAE with it, denoted as \emph{Hu-QMIX}, on the four super hard maps in SMAC. The results show that our SMMAE, which is based on the vanilla QMIX outperforms the finetuned QMIX \emph{Hu-QMIX} on all the maps~(Figure~\ref{fig:appendix_smac_vs_hujian}). Compared with \emph{Hu-QMIX}, SMMAE is able to achieve higher test win rate faster and converge to higher win rate on all maps. It demonstrates the effectiveness of SMMAE.

\subsection{Efficient Exploration in Task-Related Space}
\label{sec:task_related_space}

In this section, we qualitatively compare the exploration efficiency of SMMAE with that of baseline EMC on \emph{6h\_vs\_8z} by visualization. Figure~\ref{fig:tsne_SMMAE_vs_EMC} shows the visual results. The \emph{left} picture and the \emph{right} figure are the coordination patterns of the two algorithms during test time. Same as the operation in Figure~\ref{fig:two_dimension_tsne_ablation_SMMAE_vs_no_exppolicy}, we sample the global state points from their training processes and use t-SNE~\cite{tsne} to visualize the points after dimensionality reduction~(\emph{middle} figure of Figure~\ref{fig:tsne_SMMAE_vs_EMC}).
The \emph{middle} figure of Figure~\ref{fig:tsne_SMMAE_vs_EMC} shows that SMMAE has a similar number of clusters as that of EMC, which means SMMAE and EMC have similar sizes of exploration areas. Same as the calculation method of visiting state entropy in Section~\ref{sec:ablation_study}, the entropy of SMMAE is $13.574$, and the entropy of EMC is $13.428$. The replays of SMMAE during test time show that the agents have learned how to cooperate in different ways to attack the enemies, where the agents have learned how to surround the enemies and how to increase the distance to the enemies to take advantage of range~(\emph{left} figure of Figure~\ref{fig:tsne_SMMAE_vs_EMC}). However, the replays of EMC during test time show that the agents only have learned how to escape~(\emph{right} figure of Figure~\ref{fig:tsne_SMMAE_vs_EMC}). The different results demonstrate that although SMMAE and EMC have similar sizes of exploration areas, SMMAE can explore more efficiently because the space it explores is more task-related. In fact, our method itself does not explicitly aim to explore the task-related space. In SMMAE, each agent uses its own exploration policy to maximize the entropy of the states it visits, so the method can more efficiently cover a wider exploration space, thus covering task-related states.

\subsection{Performance on Level Based Foraging and Application on VDN}
\begin{figure}[t]
    \centering
    \begin{minipage}{\linewidth}
        \begin{minipage}[c]{0.49\linewidth}
            \centering 
            \subfigure[c][lbforaging-8x8-2p-2f]{
              \includegraphics[width=\textwidth]{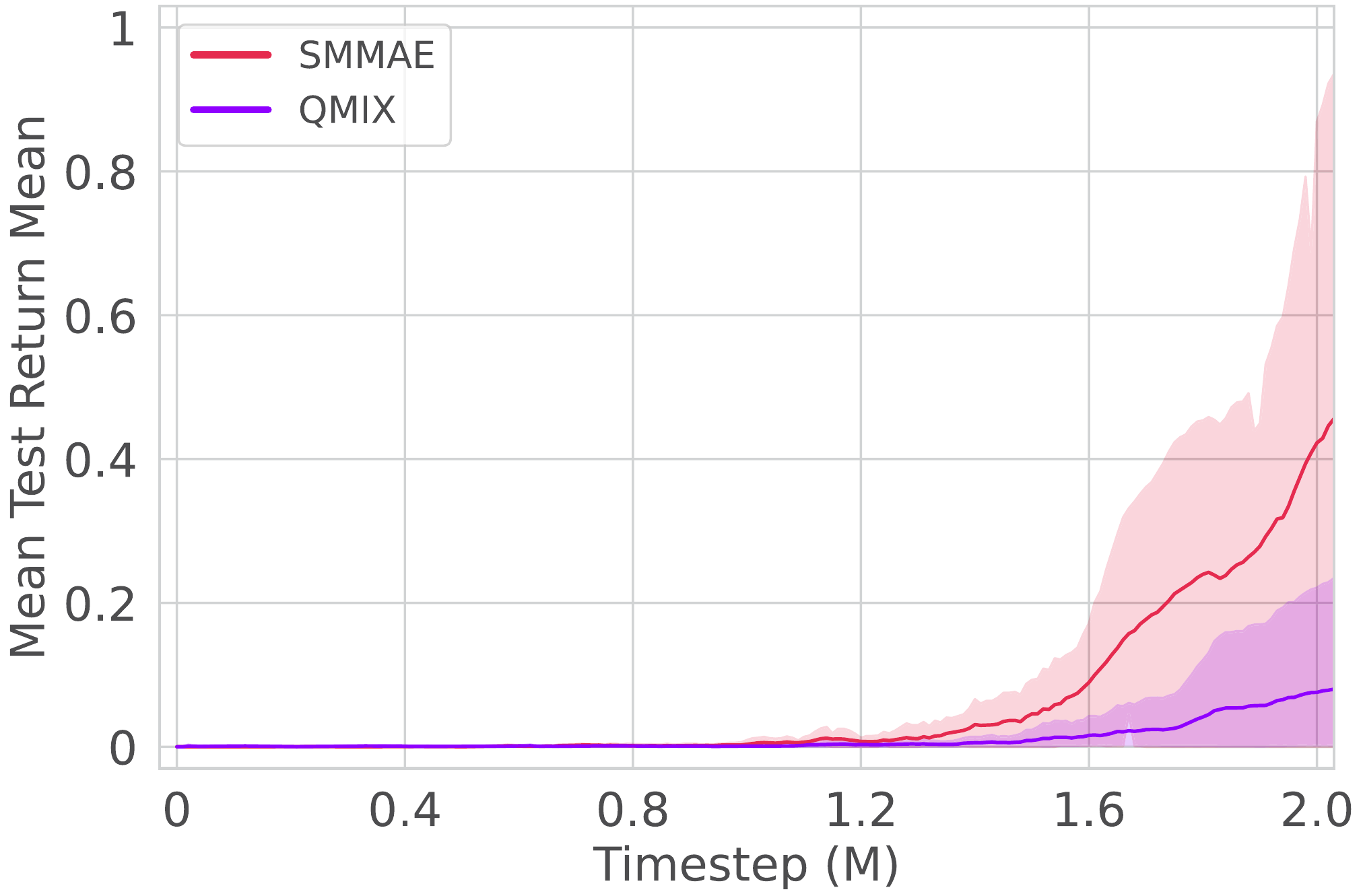}
            \label{fig:performance_lbf_8x8_2p_2f}}
        \end{minipage} 
        \begin{minipage}[c]{0.49\linewidth}
            \centering 
            \subfigure[c][lbforaging-10x10-2p-2f]{
         \includegraphics[width=\textwidth]{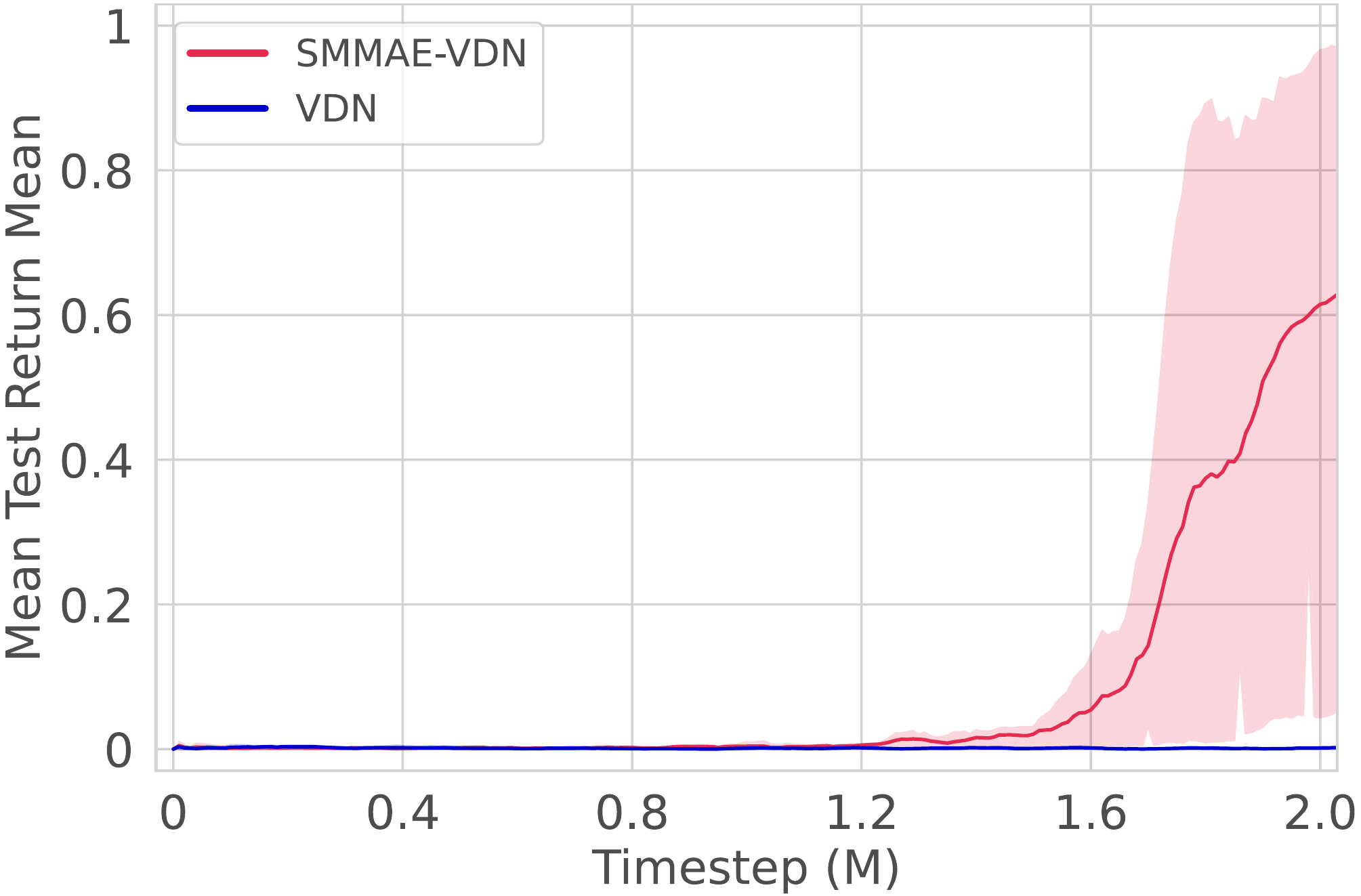}
            \label{fig:performance_lbf_10x10_2p_2f}}
        \end{minipage} 
        \vspace{-0.1in}
        \caption{Results on two scenarios in LBF.}
        \label{fig:appendix_performance_on_lbf}
    \end{minipage}
    \vspace{-0.15in}
\end{figure}

Besides SMAC, we also use Level Based Foraging~(LBF)~\cite{lbf} to test the ability of SMMAE~(Figure~\ref{fig:appendix_performance_on_lbf}). The results show that SMMAE can outperform QMIX on the simple task~(Figure~\ref{fig:performance_lbf_8x8_2p_2f}). Because VDN performs better than QMIX on LBF using PyMARL, we also apply SMMAE to another baseline VDN, denoted as SMMAE-VDN, to test the ability of SMMAE. The results show that SMMAE-VDN can also outperform VDN on the task in LBF~(Figure~\ref{fig:performance_lbf_10x10_2p_2f}). It demonstrates that SMMAE is a effective general framework, which can be applied on other value-based MARL methods, and the faster convergence speed means the baselines using SMMAE can explore more effectively.


\section{Conclusion}
\label{sec:conclusion}
With appropriate hyper-parameters, we empirically find that vanilla MARL algorithms using self-exploration can also achieve competitive performance. Consequently, in this paper we propose SMMAE, a novel algorithm that adaptively adjusts the individual exploration probability according to the uncertainty of the multi-agent system at different timesteps. SMMAE focuses on each agent's individual exploration ability by learning an individual exploration policy, which is optimized by target state distribution matching. We study SMMAE on a variety of tasks in the SMAC benchmark, and empirically reveal that SMMAE can explore more efficiently on task-related states and generate better cooperation policies. We take a step towards achieving a trade-off between individual exploration and team cooperation, and we think it is promising for SMMAE to solve complex tasks in multi-agent systems combined with other methods like communication in future work. It should be pointed out that directly matching the explored states to a uniform distribution may cause unexpected exploration, and the adaptive exploration probability is necessary for stabilizing training. It deserves further research on the target distribution of the exploration policy and the adaptive exploration probability.

\section{Acknowledgments}
\vspace{-0.03in}
This work is supported by National Key Research and Development Program of China (2020AAA0107200), the National Science Foundation of China (62250069, 61921006), and the program B for Outstanding Ph.D. candidate of Nanjing University. We would like to thank Shenghua Wan, and the anonymous reviewers for their helpful discussions and support. 




\vspace{-0.1in}
\bibliographystyle{ACM-Reference-Format} 
\bibliography{references}


\begin{thebibliography}{54}


\ifx \showCODEN    \undefined \def \showCODEN     #1{\unskip}     \fi
\ifx \showDOI      \undefined \def \showDOI       #1{#1}\fi
\ifx \showISBNx    \undefined \def \showISBNx     #1{\unskip}     \fi
\ifx \showISBNxiii \undefined \def \showISBNxiii  #1{\unskip}     \fi
\ifx \showISSN     \undefined \def \showISSN      #1{\unskip}     \fi
\ifx \showLCCN     \undefined \def \showLCCN      #1{\unskip}     \fi
\ifx \shownote     \undefined \def \shownote      #1{#1}          \fi
\ifx \showarticletitle \undefined \def \showarticletitle #1{#1}   \fi
\ifx \showURL      \undefined \def \showURL       {\relax}        \fi
\providecommand\bibfield[2]{#2}
\providecommand\bibinfo[2]{#2}
\providecommand\natexlab[1]{#1}
\providecommand\showeprint[2][]{arXiv:#2}

\bibitem[\protect\citeauthoryear{Badia, Piot, Kapturowski, Sprechmann,
  Vitvitskyi, Guo, and Blundell}{Badia et~al\mbox{.}}{2020a}]%
        {agent57}
\bibfield{author}{\bibinfo{person}{Adri{\`a}~Puigdom{\`e}nech Badia},
  \bibinfo{person}{Bilal Piot}, \bibinfo{person}{Steven Kapturowski},
  \bibinfo{person}{Pablo Sprechmann}, \bibinfo{person}{Alex Vitvitskyi},
  \bibinfo{person}{Zhaohan~Daniel Guo}, {and} \bibinfo{person}{Charles
  Blundell}.} \bibinfo{year}{2020}\natexlab{a}.
\newblock \showarticletitle{Agent57: Outperforming the atari human benchmark}.
  In \bibinfo{booktitle}{\emph{ICML}}. \bibinfo{pages}{507--517}.
\newblock


\bibitem[\protect\citeauthoryear{Badia, Sprechmann, Vitvitskyi, Guo, Piot,
  Kapturowski, Tieleman, Arjovsky, Pritzel, Bolt, et~al\mbox{.}}{Badia
  et~al\mbox{.}}{2020b}]%
        {ngu}
\bibfield{author}{\bibinfo{person}{Adri{\`a}~Puigdom{\`e}nech Badia},
  \bibinfo{person}{Pablo Sprechmann}, \bibinfo{person}{Alex Vitvitskyi},
  \bibinfo{person}{Daniel Guo}, \bibinfo{person}{Bilal Piot},
  \bibinfo{person}{Steven Kapturowski}, \bibinfo{person}{Olivier Tieleman},
  \bibinfo{person}{Mart{\'\i}n Arjovsky}, \bibinfo{person}{Alexander Pritzel},
  \bibinfo{person}{Andew Bolt}, {et~al\mbox{.}}}
  \bibinfo{year}{2020}\natexlab{b}.
\newblock \showarticletitle{Never give up: Learning directed exploration
  strategies}. In \bibinfo{booktitle}{\emph{ICLR}}.
\newblock


\bibitem[\protect\citeauthoryear{Bellemare, Srinivasan, Ostrovski, Schaul,
  Saxton, and Munos}{Bellemare et~al\mbox{.}}{2016}]%
        {unifying_cb_exploration}
\bibfield{author}{\bibinfo{person}{Marc Bellemare}, \bibinfo{person}{Sriram
  Srinivasan}, \bibinfo{person}{Georg Ostrovski}, \bibinfo{person}{Tom Schaul},
  \bibinfo{person}{David Saxton}, {and} \bibinfo{person}{Remi Munos}.}
  \bibinfo{year}{2016}\natexlab{}.
\newblock \showarticletitle{Unifying count-based exploration and intrinsic
  motivation}. In \bibinfo{booktitle}{\emph{NeurIPS}}.
\newblock


\bibitem[\protect\citeauthoryear{Berseth, Geng, Devin, Rhinehart, Finn,
  Jayaraman, and Levine}{Berseth et~al\mbox{.}}{2021}]%
        {smirl}
\bibfield{author}{\bibinfo{person}{Glen Berseth}, \bibinfo{person}{Daniel
  Geng}, \bibinfo{person}{Coline Devin}, \bibinfo{person}{Nicholas Rhinehart},
  \bibinfo{person}{Chelsea Finn}, \bibinfo{person}{Dinesh Jayaraman}, {and}
  \bibinfo{person}{Sergey Levine}.} \bibinfo{year}{2021}\natexlab{}.
\newblock \showarticletitle{SMiRL: Surprise Minimizing Reinforcement Learning
  in Unstable Environments}. In \bibinfo{booktitle}{\emph{ICLR}}.
\newblock


\bibitem[\protect\citeauthoryear{Burda, Edwards, Storkey, and Klimov}{Burda
  et~al\mbox{.}}{2019}]%
        {rnd}
\bibfield{author}{\bibinfo{person}{Yuri Burda}, \bibinfo{person}{Harrison
  Edwards}, \bibinfo{person}{Amos Storkey}, {and} \bibinfo{person}{Oleg
  Klimov}.} \bibinfo{year}{2019}\natexlab{}.
\newblock \showarticletitle{Exploration by random network distillation}. In
  \bibinfo{booktitle}{\emph{ICLR}}.
\newblock


\bibitem[\protect\citeauthoryear{Cao, Yuan, Wang, Zhang, Zhang, Yu, and
  Zhan}{Cao et~al\mbox{.}}{2021}]%
        {cao2021linda}
\bibfield{author}{\bibinfo{person}{Jiahan Cao}, \bibinfo{person}{Lei Yuan},
  \bibinfo{person}{Jianhao Wang}, \bibinfo{person}{Shaowei Zhang},
  \bibinfo{person}{Chongjie Zhang}, \bibinfo{person}{Yang Yu}, {and}
  \bibinfo{person}{De-Chuan Zhan}.} \bibinfo{year}{2021}\natexlab{}.
\newblock \showarticletitle{LINDA: Multi-Agent Local Information Decomposition
  for Awareness of Teammates}.
\newblock \bibinfo{journal}{\emph{arXiv preprint arXiv:2109.12508}}
  (\bibinfo{year}{2021}).
\newblock


\bibitem[\protect\citeauthoryear{Christianos, Sch{\"a}fer, and
  Albrecht}{Christianos et~al\mbox{.}}{2020}]%
        {lbf}
\bibfield{author}{\bibinfo{person}{Filippos Christianos},
  \bibinfo{person}{Lukas Sch{\"a}fer}, {and} \bibinfo{person}{Stefano
  Albrecht}.} \bibinfo{year}{2020}\natexlab{}.
\newblock \showarticletitle{Shared experience actor-critic for multi-agent
  reinforcement learning}.
\newblock \bibinfo{journal}{\emph{NeurIPS}} (\bibinfo{year}{2020}).
\newblock


\bibitem[\protect\citeauthoryear{Dabney, Ostrovski, and Barreto}{Dabney
  et~al\mbox{.}}{2021}]%
        {epsilonzgreedy}
\bibfield{author}{\bibinfo{person}{Will Dabney}, \bibinfo{person}{Georg
  Ostrovski}, {and} \bibinfo{person}{Andr{\'{e}} Barreto}.}
  \bibinfo{year}{2021}\natexlab{}.
\newblock \showarticletitle{Temporally-Extended {\(\epsilon\)}-Greedy
  Exploration}. In \bibinfo{booktitle}{\emph{ICLR}}.
\newblock


\bibitem[\protect\citeauthoryear{Eysenbach, Gupta, Ibarz, and Levine}{Eysenbach
  et~al\mbox{.}}{2019}]%
        {diversity_is_all_you_need}
\bibfield{author}{\bibinfo{person}{Benjamin Eysenbach},
  \bibinfo{person}{Abhishek Gupta}, \bibinfo{person}{Julian Ibarz}, {and}
  \bibinfo{person}{Sergey Levine}.} \bibinfo{year}{2019}\natexlab{}.
\newblock \showarticletitle{Diversity is all you need: Learning skills without
  a reward function}. In \bibinfo{booktitle}{\emph{ICLR}}.
\newblock


\bibitem[\protect\citeauthoryear{Foerster, Farquhar, Afouras, Nardelli, and
  Whiteson}{Foerster et~al\mbox{.}}{2018}]%
        {coma}
\bibfield{author}{\bibinfo{person}{Jakob Foerster}, \bibinfo{person}{Gregory
  Farquhar}, \bibinfo{person}{Triantafyllos Afouras}, \bibinfo{person}{Nantas
  Nardelli}, {and} \bibinfo{person}{Shimon Whiteson}.}
  \bibinfo{year}{2018}\natexlab{}.
\newblock \showarticletitle{Counterfactual multi-agent policy gradients}. In
  \bibinfo{booktitle}{\emph{AAAI}}.
\newblock


\bibitem[\protect\citeauthoryear{Fortunato, Azar, Piot, Menick, Hessel, Osband,
  Graves, Mnih, Munos, Hassabis, Pietquin, Blundell, and Legg}{Fortunato
  et~al\mbox{.}}{2018}]%
        {noisynet}
\bibfield{author}{\bibinfo{person}{Meire Fortunato},
  \bibinfo{person}{Mohammad~Gheshlaghi Azar}, \bibinfo{person}{Bilal Piot},
  \bibinfo{person}{Jacob Menick}, \bibinfo{person}{Matteo Hessel},
  \bibinfo{person}{Ian Osband}, \bibinfo{person}{Alex Graves},
  \bibinfo{person}{Volodymyr Mnih}, \bibinfo{person}{R{\'{e}}mi Munos},
  \bibinfo{person}{Demis Hassabis}, \bibinfo{person}{Olivier Pietquin},
  \bibinfo{person}{Charles Blundell}, {and} \bibinfo{person}{Shane Legg}.}
  \bibinfo{year}{2018}\natexlab{}.
\newblock \showarticletitle{Noisy Networks For Exploration}. In
  \bibinfo{booktitle}{\emph{ICLR}}.
\newblock


\bibitem[\protect\citeauthoryear{Greshler, Gordon, Salzman, and
  Shimkin}{Greshler et~al\mbox{.}}{2021}]%
        {greshler2021cooperative}
\bibfield{author}{\bibinfo{person}{Nir Greshler}, \bibinfo{person}{Ofir
  Gordon}, \bibinfo{person}{Oren Salzman}, {and} \bibinfo{person}{Nahum
  Shimkin}.} \bibinfo{year}{2021}\natexlab{}.
\newblock \showarticletitle{Cooperative Multi-Agent Path Finding: Beyond Path
  Planning and Collision Avoidance}. In \bibinfo{booktitle}{\emph{MRS}}.
  \bibinfo{pages}{20--28}.
\newblock


\bibitem[\protect\citeauthoryear{Gronauer and Diepold}{Gronauer and
  Diepold}{2022}]%
        {gronauer2022multi}
\bibfield{author}{\bibinfo{person}{Sven Gronauer} {and} \bibinfo{person}{Klaus
  Diepold}.} \bibinfo{year}{2022}\natexlab{}.
\newblock \showarticletitle{Multi-agent deep reinforcement learning: a survey}.
\newblock \bibinfo{journal}{\emph{Artificial Intelligence Review}}
  \bibinfo{volume}{55}, \bibinfo{number}{2} (\bibinfo{year}{2022}),
  \bibinfo{pages}{895--943}.
\newblock


\bibitem[\protect\citeauthoryear{Hu, Jiang, Harding, Wu, and Liao}{Hu
  et~al\mbox{.}}{2021}]%
        {rethinkingqmix}
\bibfield{author}{\bibinfo{person}{Jian Hu}, \bibinfo{person}{Siyang Jiang},
  \bibinfo{person}{Seth~Austin Harding}, \bibinfo{person}{Haibin Wu}, {and}
  \bibinfo{person}{Shih-wei Liao}.} \bibinfo{year}{2021}\natexlab{}.
\newblock \showarticletitle{Rethinking the implementation tricks and
  monotonicity constraint in cooperative multi-agent reinforcement learning}.
\newblock \bibinfo{journal}{\emph{CoRR}} (\bibinfo{year}{2021}).
\newblock


\bibitem[\protect\citeauthoryear{Kingma and Welling}{Kingma and
  Welling}{2014}]%
        {vae}
\bibfield{author}{\bibinfo{person}{Diederik~P. Kingma} {and}
  \bibinfo{person}{Max Welling}.} \bibinfo{year}{2014}\natexlab{}.
\newblock \showarticletitle{Auto-Encoding Variational Bayes}. In
  \bibinfo{booktitle}{\emph{ICLR}}.
\newblock


\bibitem[\protect\citeauthoryear{Klein, Orasanu, Calderwood, Zsambok,
  et~al\mbox{.}}{Klein et~al\mbox{.}}{1993}]%
        {klein1993decision}
\bibfield{author}{\bibinfo{person}{Gary~A Klein}, \bibinfo{person}{Judith
  Orasanu}, \bibinfo{person}{Roberta Calderwood}, \bibinfo{person}{Caroline~E
  Zsambok}, {et~al\mbox{.}}} \bibinfo{year}{1993}\natexlab{}.
\newblock \bibinfo{booktitle}{\emph{Decision Making in Action: Models and
  Methods}}.
\newblock \bibinfo{publisher}{Ablex Norwood, NJ}.
\newblock


\bibitem[\protect\citeauthoryear{Kraskov, St{\"o}gbauer, and
  Grassberger}{Kraskov et~al\mbox{.}}{2004}]%
        {kraskov2004estimating}
\bibfield{author}{\bibinfo{person}{Alexander Kraskov}, \bibinfo{person}{Harald
  St{\"o}gbauer}, {and} \bibinfo{person}{Peter Grassberger}.}
  \bibinfo{year}{2004}\natexlab{}.
\newblock \showarticletitle{Estimating mutual information}.
\newblock \bibinfo{journal}{\emph{Physical review E}} \bibinfo{volume}{69},
  \bibinfo{number}{6} (\bibinfo{year}{2004}), \bibinfo{pages}{066138}.
\newblock


\bibitem[\protect\citeauthoryear{Lee, Eysenbach, Parisotto, Xing, Levine, and
  Salakhutdinov}{Lee et~al\mbox{.}}{2019}]%
        {smm}
\bibfield{author}{\bibinfo{person}{Lisa Lee}, \bibinfo{person}{Benjamin
  Eysenbach}, \bibinfo{person}{Emilio Parisotto}, \bibinfo{person}{Eric Xing},
  \bibinfo{person}{Sergey Levine}, {and} \bibinfo{person}{Ruslan
  Salakhutdinov}.} \bibinfo{year}{2019}\natexlab{}.
\newblock \showarticletitle{Efficient exploration via state marginal matching}.
\newblock \bibinfo{journal}{\emph{CoRR}}  \bibinfo{volume}{abs/1906.05274}
  (\bibinfo{year}{2019}).
\newblock


\bibitem[\protect\citeauthoryear{Liu, Jain, Yeh, and Schwing}{Liu
  et~al\mbox{.}}{2021}]%
        {cmae}
\bibfield{author}{\bibinfo{person}{Iou-Jen Liu}, \bibinfo{person}{Unnat Jain},
  \bibinfo{person}{Raymond~A Yeh}, {and} \bibinfo{person}{Alexander Schwing}.}
  \bibinfo{year}{2021}\natexlab{}.
\newblock \showarticletitle{Cooperative exploration for multi-agent deep
  reinforcement learning}. In \bibinfo{booktitle}{\emph{ICML}}.
  \bibinfo{pages}{6826--6836}.
\newblock


\bibitem[\protect\citeauthoryear{Lowe, Wu, Tamar, Harb, Pieter~Abbeel, and
  Mordatch}{Lowe et~al\mbox{.}}{2017}]%
        {maddpg}
\bibfield{author}{\bibinfo{person}{Ryan Lowe}, \bibinfo{person}{Yi~I Wu},
  \bibinfo{person}{Aviv Tamar}, \bibinfo{person}{Jean Harb},
  \bibinfo{person}{OpenAI Pieter~Abbeel}, {and} \bibinfo{person}{Igor
  Mordatch}.} \bibinfo{year}{2017}\natexlab{}.
\newblock \showarticletitle{Multi-agent actor-critic for mixed
  cooperative-competitive environments}. In
  \bibinfo{booktitle}{\emph{NeurIPS}}.
\newblock


\bibitem[\protect\citeauthoryear{Mahajan, Rashid, Samvelyan, and
  Whiteson}{Mahajan et~al\mbox{.}}{2019}]%
        {maven}
\bibfield{author}{\bibinfo{person}{Anuj Mahajan}, \bibinfo{person}{Tabish
  Rashid}, \bibinfo{person}{Mikayel Samvelyan}, {and} \bibinfo{person}{Shimon
  Whiteson}.} \bibinfo{year}{2019}\natexlab{}.
\newblock \showarticletitle{Maven: Multi-agent variational exploration}.
\newblock \bibinfo{journal}{\emph{NeurIPS}} (\bibinfo{year}{2019}).
\newblock


\bibitem[\protect\citeauthoryear{Mnih, Kavukcuoglu, Silver, Rusu, Veness,
  Bellemare, Graves, Riedmiller, Fidjeland, Ostrovski, et~al\mbox{.}}{Mnih
  et~al\mbox{.}}{2015}]%
        {dqn_nature}
\bibfield{author}{\bibinfo{person}{Volodymyr Mnih}, \bibinfo{person}{Koray
  Kavukcuoglu}, \bibinfo{person}{David Silver}, \bibinfo{person}{Andrei~A
  Rusu}, \bibinfo{person}{Joel Veness}, \bibinfo{person}{Marc~G Bellemare},
  \bibinfo{person}{Alex Graves}, \bibinfo{person}{Martin Riedmiller},
  \bibinfo{person}{Andreas~K Fidjeland}, \bibinfo{person}{Georg Ostrovski},
  {et~al\mbox{.}}} \bibinfo{year}{2015}\natexlab{}.
\newblock \showarticletitle{Human-level control through deep reinforcement
  learning}.
\newblock \bibinfo{journal}{\emph{nature}} \bibinfo{volume}{518},
  \bibinfo{number}{7540} (\bibinfo{year}{2015}), \bibinfo{pages}{529--533}.
\newblock


\bibitem[\protect\citeauthoryear{Oliehoek and Amato}{Oliehoek and
  Amato}{2016}]%
        {decpomdp}
\bibfield{author}{\bibinfo{person}{Frans~A Oliehoek} {and}
  \bibinfo{person}{Christopher Amato}.} \bibinfo{year}{2016}\natexlab{}.
\newblock \bibinfo{booktitle}{\emph{A Concise Introduction to Decentralized
  POMDPs}}.
\newblock \bibinfo{publisher}{Springer}.
\newblock


\bibitem[\protect\citeauthoryear{Oliehoek, Spaan, and Vlassis}{Oliehoek
  et~al\mbox{.}}{2008}]%
        {ctde_1}
\bibfield{author}{\bibinfo{person}{Frans~A Oliehoek},
  \bibinfo{person}{Matthijs~TJ Spaan}, {and} \bibinfo{person}{Nikos Vlassis}.}
  \bibinfo{year}{2008}\natexlab{}.
\newblock \showarticletitle{Optimal and approximate Q-value functions for
  decentralized POMDPs}.
\newblock \bibinfo{journal}{\emph{Journal of Artificial Intelligence Research}}
   \bibinfo{volume}{32} (\bibinfo{year}{2008}), \bibinfo{pages}{289--353}.
\newblock


\bibitem[\protect\citeauthoryear{OroojlooyJadid and Hajinezhad}{OroojlooyJadid
  and Hajinezhad}{2019}]%
        {oroojlooyjadid2019review}
\bibfield{author}{\bibinfo{person}{Afshin OroojlooyJadid} {and}
  \bibinfo{person}{Davood Hajinezhad}.} \bibinfo{year}{2019}\natexlab{}.
\newblock \showarticletitle{A review of cooperative multi-agent deep
  reinforcement learning}.
\newblock \bibinfo{journal}{\emph{arXiv preprint arXiv:1908.03963}}
  (\bibinfo{year}{2019}).
\newblock


\bibitem[\protect\citeauthoryear{Ostrovski, Bellemare, Oord, and
  Munos}{Ostrovski et~al\mbox{.}}{2017}]%
        {ostrovski2017count}
\bibfield{author}{\bibinfo{person}{Georg Ostrovski}, \bibinfo{person}{Marc~G
  Bellemare}, \bibinfo{person}{A{\"a}ron Oord}, {and} \bibinfo{person}{R{\'e}mi
  Munos}.} \bibinfo{year}{2017}\natexlab{}.
\newblock \showarticletitle{Count-based exploration with neural density
  models}. In \bibinfo{booktitle}{\emph{ICML}}. \bibinfo{pages}{2721--2730}.
\newblock


\bibitem[\protect\citeauthoryear{Parisi, Dean, Pathak, and Gupta}{Parisi
  et~al\mbox{.}}{2021}]%
        {c_bet}
\bibfield{author}{\bibinfo{person}{Simone Parisi}, \bibinfo{person}{Victoria
  Dean}, \bibinfo{person}{Deepak Pathak}, {and} \bibinfo{person}{Abhinav
  Gupta}.} \bibinfo{year}{2021}\natexlab{}.
\newblock \showarticletitle{Interesting Object, Curious Agent: Learning
  Task-Agnostic Exploration}.
\newblock \bibinfo{journal}{\emph{NeurIPS}} (\bibinfo{year}{2021}).
\newblock


\bibitem[\protect\citeauthoryear{Pathak, Agrawal, Efros, and Darrell}{Pathak
  et~al\mbox{.}}{2017}]%
        {icm}
\bibfield{author}{\bibinfo{person}{Deepak Pathak}, \bibinfo{person}{Pulkit
  Agrawal}, \bibinfo{person}{Alexei~A Efros}, {and} \bibinfo{person}{Trevor
  Darrell}.} \bibinfo{year}{2017}\natexlab{}.
\newblock \showarticletitle{Curiosity-driven exploration by self-supervised
  prediction}. In \bibinfo{booktitle}{\emph{ICML}}.
  \bibinfo{pages}{2778--2787}.
\newblock


\bibitem[\protect\citeauthoryear{Pathak, Gandhi, and Gupta}{Pathak
  et~al\mbox{.}}{2019}]%
        {exploration_via_disagreement}
\bibfield{author}{\bibinfo{person}{Deepak Pathak}, \bibinfo{person}{Dhiraj
  Gandhi}, {and} \bibinfo{person}{Abhinav Gupta}.}
  \bibinfo{year}{2019}\natexlab{}.
\newblock \showarticletitle{Self-supervised exploration via disagreement}. In
  \bibinfo{booktitle}{\emph{ICML}}. \bibinfo{pages}{5062--5071}.
\newblock


\bibitem[\protect\citeauthoryear{P{\^\i}slar, Szepesvari, Ostrovski, Borsa, and
  Schaul}{P{\^\i}slar et~al\mbox{.}}{2022}]%
        {when_should_agents_explore}
\bibfield{author}{\bibinfo{person}{Miruna P{\^\i}slar}, \bibinfo{person}{David
  Szepesvari}, \bibinfo{person}{Georg Ostrovski}, \bibinfo{person}{Diana
  Borsa}, {and} \bibinfo{person}{Tom Schaul}.} \bibinfo{year}{2022}\natexlab{}.
\newblock \showarticletitle{When should agents explore?}. In
  \bibinfo{booktitle}{\emph{ICLR}}.
\newblock


\bibitem[\protect\citeauthoryear{Pitis, Chan, Zhao, Stadie, and Ba}{Pitis
  et~al\mbox{.}}{2020}]%
        {mega_omega}
\bibfield{author}{\bibinfo{person}{Silviu Pitis}, \bibinfo{person}{Harris
  Chan}, \bibinfo{person}{Stephen Zhao}, \bibinfo{person}{Bradly Stadie}, {and}
  \bibinfo{person}{Jimmy Ba}.} \bibinfo{year}{2020}\natexlab{}.
\newblock \showarticletitle{Maximum entropy gain exploration for long horizon
  multi-goal reinforcement learning}. In \bibinfo{booktitle}{\emph{ICML}}.
  \bibinfo{pages}{7750--7761}.
\newblock


\bibitem[\protect\citeauthoryear{Plappert, Houthooft, Dhariwal, Sidor, Chen,
  Chen, Asfour, Abbeel, and Andrychowicz}{Plappert et~al\mbox{.}}{2018}]%
        {parameter_space_noise}
\bibfield{author}{\bibinfo{person}{Matthias Plappert}, \bibinfo{person}{Rein
  Houthooft}, \bibinfo{person}{Prafulla Dhariwal}, \bibinfo{person}{Szymon
  Sidor}, \bibinfo{person}{Richard~Y. Chen}, \bibinfo{person}{Xi Chen},
  \bibinfo{person}{Tamim Asfour}, \bibinfo{person}{Pieter Abbeel}, {and}
  \bibinfo{person}{Marcin Andrychowicz}.} \bibinfo{year}{2018}\natexlab{}.
\newblock \showarticletitle{Parameter Space Noise for Exploration}. In
  \bibinfo{booktitle}{\emph{ICLR}}.
\newblock


\bibitem[\protect\citeauthoryear{Rashid, Farquhar, Peng, and Whiteson}{Rashid
  et~al\mbox{.}}{2020}]%
        {wqmix}
\bibfield{author}{\bibinfo{person}{Tabish Rashid}, \bibinfo{person}{Gregory
  Farquhar}, \bibinfo{person}{Bei Peng}, {and} \bibinfo{person}{Shimon
  Whiteson}.} \bibinfo{year}{2020}\natexlab{}.
\newblock \showarticletitle{Weighted qmix: Expanding monotonic value function
  factorisation for deep multi-agent reinforcement learning}.
\newblock \bibinfo{journal}{\emph{NeurIPS}} (\bibinfo{year}{2020}),
  \bibinfo{pages}{10199--10210}.
\newblock


\bibitem[\protect\citeauthoryear{Rashid, Samvelyan, Schroeder, Farquhar,
  Foerster, and Whiteson}{Rashid et~al\mbox{.}}{2018}]%
        {qmix}
\bibfield{author}{\bibinfo{person}{Tabish Rashid}, \bibinfo{person}{Mikayel
  Samvelyan}, \bibinfo{person}{Christian Schroeder}, \bibinfo{person}{Gregory
  Farquhar}, \bibinfo{person}{Jakob Foerster}, {and} \bibinfo{person}{Shimon
  Whiteson}.} \bibinfo{year}{2018}\natexlab{}.
\newblock \showarticletitle{Qmix: Monotonic value function factorisation for
  deep multi-agent reinforcement learning}. In
  \bibinfo{booktitle}{\emph{ICML}}. \bibinfo{pages}{4295--4304}.
\newblock


\bibitem[\protect\citeauthoryear{Samvelyan, Rashid, De~Witt, Farquhar,
  Nardelli, Rudner, Hung, Torr, Foerster, and Whiteson}{Samvelyan
  et~al\mbox{.}}{2019}]%
        {smac}
\bibfield{author}{\bibinfo{person}{Mikayel Samvelyan}, \bibinfo{person}{Tabish
  Rashid}, \bibinfo{person}{Christian~Schroeder De~Witt},
  \bibinfo{person}{Gregory Farquhar}, \bibinfo{person}{Nantas Nardelli},
  \bibinfo{person}{Tim~GJ Rudner}, \bibinfo{person}{Chia-Man Hung},
  \bibinfo{person}{Philip~HS Torr}, \bibinfo{person}{Jakob Foerster}, {and}
  \bibinfo{person}{Shimon Whiteson}.} \bibinfo{year}{2019}\natexlab{}.
\newblock \showarticletitle{The starcraft multi-agent challenge}. In
  \bibinfo{booktitle}{\emph{AAMAS}}. \bibinfo{pages}{2186--2188}.
\newblock


\bibitem[\protect\citeauthoryear{Sunehag, Lever, Gruslys, Czarnecki, Zambaldi,
  Jaderberg, Lanctot, Sonnerat, Leibo, Tuyls, et~al\mbox{.}}{Sunehag
  et~al\mbox{.}}{2018}]%
        {vdn}
\bibfield{author}{\bibinfo{person}{Peter Sunehag}, \bibinfo{person}{Guy Lever},
  \bibinfo{person}{Audrunas Gruslys}, \bibinfo{person}{Wojciech~Marian
  Czarnecki}, \bibinfo{person}{Vinicius Zambaldi}, \bibinfo{person}{Max
  Jaderberg}, \bibinfo{person}{Marc Lanctot}, \bibinfo{person}{Nicolas
  Sonnerat}, \bibinfo{person}{Joel~Z Leibo}, \bibinfo{person}{Karl Tuyls},
  {et~al\mbox{.}}} \bibinfo{year}{2018}\natexlab{}.
\newblock \showarticletitle{Value-decomposition networks for cooperative
  multi-agent learning}. In \bibinfo{booktitle}{\emph{AAMAS}}.
  \bibinfo{pages}{2085--2087}.
\newblock


\bibitem[\protect\citeauthoryear{Sutton and Barto}{Sutton and Barto}{2018}]%
        {sutton2018reinforcement}
\bibfield{author}{\bibinfo{person}{Richard~S Sutton} {and}
  \bibinfo{person}{Andrew~G Barto}.} \bibinfo{year}{2018}\natexlab{}.
\newblock \bibinfo{booktitle}{\emph{Reinforcement learning: An introduction}}.
\newblock \bibinfo{publisher}{MIT press}.
\newblock


\bibitem[\protect\citeauthoryear{Tokic}{Tokic}{2010}]%
        {adaptiveepsilongreedy}
\bibfield{author}{\bibinfo{person}{Michel Tokic}.}
  \bibinfo{year}{2010}\natexlab{}.
\newblock \showarticletitle{Adaptive $\varepsilon$-greedy exploration in
  reinforcement learning based on value differences}. In
  \bibinfo{booktitle}{\emph{AAAI}}. \bibinfo{pages}{203--210}.
\newblock


\bibitem[\protect\citeauthoryear{Van~der Maaten and Hinton}{Van~der Maaten and
  Hinton}{2008}]%
        {tsne}
\bibfield{author}{\bibinfo{person}{Laurens Van~der Maaten} {and}
  \bibinfo{person}{Geoffrey Hinton}.} \bibinfo{year}{2008}\natexlab{}.
\newblock \showarticletitle{Visualizing data using t-{SNE}}.
\newblock \bibinfo{journal}{\emph{Journal of Machine Learning Research}}
  \bibinfo{volume}{9}, \bibinfo{number}{11} (\bibinfo{year}{2008}),
  \bibinfo{pages}{2579--2605}.
\newblock


\bibitem[\protect\citeauthoryear{Van~Hasselt, Guez, and Silver}{Van~Hasselt
  et~al\mbox{.}}{2016}]%
        {double_dqn}
\bibfield{author}{\bibinfo{person}{Hado Van~Hasselt}, \bibinfo{person}{Arthur
  Guez}, {and} \bibinfo{person}{David Silver}.}
  \bibinfo{year}{2016}\natexlab{}.
\newblock \showarticletitle{Deep reinforcement learning with double
  q-learning}. In \bibinfo{booktitle}{\emph{AAAI}}.
\newblock


\bibitem[\protect\citeauthoryear{Vaswani, Shazeer, Parmar, Uszkoreit, Jones,
  Gomez, Kaiser, and Polosukhin}{Vaswani et~al\mbox{.}}{2017}]%
        {transformer}
\bibfield{author}{\bibinfo{person}{Ashish Vaswani}, \bibinfo{person}{Noam
  Shazeer}, \bibinfo{person}{Niki Parmar}, \bibinfo{person}{Jakob Uszkoreit},
  \bibinfo{person}{Llion Jones}, \bibinfo{person}{Aidan~N Gomez},
  \bibinfo{person}{{\L}ukasz Kaiser}, {and} \bibinfo{person}{Illia
  Polosukhin}.} \bibinfo{year}{2017}\natexlab{}.
\newblock \showarticletitle{Attention is all you need}. In
  \bibinfo{booktitle}{\emph{NeurIPS}}.
\newblock


\bibitem[\protect\citeauthoryear{Wang and Wong}{Wang and Wong}{2021}]%
        {marl_nlp}
\bibfield{author}{\bibinfo{person}{Huimin Wang} {and} \bibinfo{person}{Kam-Fai
  Wong}.} \bibinfo{year}{2021}\natexlab{}.
\newblock \showarticletitle{A Collaborative Multi-agent Reinforcement Learning
  Framework for Dialog Action Decomposition}. In
  \bibinfo{booktitle}{\emph{EMNLP}}. \bibinfo{pages}{7882--7889}.
\newblock


\bibitem[\protect\citeauthoryear{Wang, Ren, Liu, Yu, and Zhang}{Wang
  et~al\mbox{.}}{2021b}]%
        {qplex}
\bibfield{author}{\bibinfo{person}{Jianhao Wang}, \bibinfo{person}{Zhizhou
  Ren}, \bibinfo{person}{Terry Liu}, \bibinfo{person}{Yang Yu}, {and}
  \bibinfo{person}{Chongjie Zhang}.} \bibinfo{year}{2021}\natexlab{b}.
\newblock \showarticletitle{{QPLEX:} Duplex Dueling Multi-Agent Q-Learning}. In
  \bibinfo{booktitle}{\emph{ICLR}}.
\newblock


\bibitem[\protect\citeauthoryear{Wang, Gupta, Mahajan, Peng, Whiteson, and
  Zhang}{Wang et~al\mbox{.}}{2021a}]%
        {rode}
\bibfield{author}{\bibinfo{person}{Tonghan Wang}, \bibinfo{person}{Tarun
  Gupta}, \bibinfo{person}{Anuj Mahajan}, \bibinfo{person}{Bei Peng},
  \bibinfo{person}{Shimon Whiteson}, {and} \bibinfo{person}{Chongjie Zhang}.}
  \bibinfo{year}{2021}\natexlab{a}.
\newblock \showarticletitle{{RODE:} Learning Roles to Decompose Multi-Agent
  Tasks}. In \bibinfo{booktitle}{\emph{ICLR}}.
\newblock


\bibitem[\protect\citeauthoryear{Wang, Wang, Wu, and Zhang}{Wang
  et~al\mbox{.}}{2020}]%
        {eiti_edti}
\bibfield{author}{\bibinfo{person}{Tonghan Wang}, \bibinfo{person}{Jianhao
  Wang}, \bibinfo{person}{Yi Wu}, {and} \bibinfo{person}{Chongjie Zhang}.}
  \bibinfo{year}{2020}\natexlab{}.
\newblock \showarticletitle{Influence-based multi-agent exploration}. In
  \bibinfo{booktitle}{\emph{ICLR}}.
\newblock


\bibitem[\protect\citeauthoryear{Wang, Schaul, Hessel, Hasselt, Lanctot, and
  Freitas}{Wang et~al\mbox{.}}{2016}]%
        {dueling_dqn}
\bibfield{author}{\bibinfo{person}{Ziyu Wang}, \bibinfo{person}{Tom Schaul},
  \bibinfo{person}{Matteo Hessel}, \bibinfo{person}{Hado Hasselt},
  \bibinfo{person}{Marc Lanctot}, {and} \bibinfo{person}{Nando Freitas}.}
  \bibinfo{year}{2016}\natexlab{}.
\newblock \showarticletitle{Dueling network architectures for deep
  reinforcement learning}. In \bibinfo{booktitle}{\emph{ICML}}.
  \bibinfo{pages}{1995--2003}.
\newblock


\bibitem[\protect\citeauthoryear{Xue, Xu, Yuan, Li, Qian, Zhang, and Yu}{Xue
  et~al\mbox{.}}{2022}]%
        {madac}
\bibfield{author}{\bibinfo{person}{Ke Xue}, \bibinfo{person}{Jiacheng Xu},
  \bibinfo{person}{Lei Yuan}, \bibinfo{person}{Miqing Li},
  \bibinfo{person}{Chao Qian}, \bibinfo{person}{Zongzhang Zhang}, {and}
  \bibinfo{person}{Yang Yu}.} \bibinfo{year}{2022}\natexlab{}.
\newblock \showarticletitle{Multi-agent Dynamic Algorithm Configuration}. In
  \bibinfo{booktitle}{\emph{NeurIPS}}.
\newblock


\bibitem[\protect\citeauthoryear{Yang, Tang, Bai, Liu, Hao, Meng, and Liu}{Yang
  et~al\mbox{.}}{2021}]%
        {exploration_survey1}
\bibfield{author}{\bibinfo{person}{Tianpei Yang}, \bibinfo{person}{Hongyao
  Tang}, \bibinfo{person}{Chenjia Bai}, \bibinfo{person}{Jinyi Liu},
  \bibinfo{person}{Jianye Hao}, \bibinfo{person}{Zhaopeng Meng}, {and}
  \bibinfo{person}{Peng Liu}.} \bibinfo{year}{2021}\natexlab{}.
\newblock \showarticletitle{Exploration in Deep Reinforcement Learning: {A}
  Comprehensive Survey}.
\newblock \bibinfo{journal}{\emph{CoRR}}  \bibinfo{volume}{abs/2109.06668}
  (\bibinfo{year}{2021}).
\newblock


\bibitem[\protect\citeauthoryear{Yu, Velu, Vinitsky, Wang, Bayen, and Wu}{Yu
  et~al\mbox{.}}{2021}]%
        {mappo}
\bibfield{author}{\bibinfo{person}{Chao Yu}, \bibinfo{person}{Akash Velu},
  \bibinfo{person}{Eugene Vinitsky}, \bibinfo{person}{Yu Wang},
  \bibinfo{person}{Alexandre Bayen}, {and} \bibinfo{person}{Yi Wu}.}
  \bibinfo{year}{2021}\natexlab{}.
\newblock \showarticletitle{The surprising effectiveness of ppo in cooperative,
  multi-agent games}.
\newblock \bibinfo{journal}{\emph{arXiv preprint arXiv:2103.01955}}
  (\bibinfo{year}{2021}).
\newblock


\bibitem[\protect\citeauthoryear{Yuan, Wang, Wang, Zhang, Chen, Guan, Zhang,
  Zhang, and Yu}{Yuan et~al\mbox{.}}{2022a}]%
        {DBLP:conf/ijcai/YuanWWZCGZZY22}
\bibfield{author}{\bibinfo{person}{Lei Yuan}, \bibinfo{person}{Chenghe Wang},
  \bibinfo{person}{Jianhao Wang}, \bibinfo{person}{Fuxiang Zhang},
  \bibinfo{person}{Feng Chen}, \bibinfo{person}{Cong Guan},
  \bibinfo{person}{Zongzhang Zhang}, \bibinfo{person}{Chongjie Zhang}, {and}
  \bibinfo{person}{Yang Yu}.} \bibinfo{year}{2022}\natexlab{a}.
\newblock \showarticletitle{Multi-Agent Concentrative Coordination with
  Decentralized Task Representation}. In \bibinfo{booktitle}{\emph{IJCAI}}.
  \bibinfo{pages}{599--605}.
\newblock


\bibitem[\protect\citeauthoryear{Yuan, Wang, Zhang, Wang, Zhang, Yu, and
  Zhang}{Yuan et~al\mbox{.}}{2022b}]%
        {DBLP:conf/aaai/YuanWZWZ0Z22}
\bibfield{author}{\bibinfo{person}{Lei Yuan}, \bibinfo{person}{Jianhao Wang},
  \bibinfo{person}{Fuxiang Zhang}, \bibinfo{person}{Chenghe Wang},
  \bibinfo{person}{Zongzhang Zhang}, \bibinfo{person}{Yang Yu}, {and}
  \bibinfo{person}{Chongjie Zhang}.} \bibinfo{year}{2022}\natexlab{b}.
\newblock \showarticletitle{Multi-Agent Incentive Communication via
  Decentralized Teammate Modeling}. In \bibinfo{booktitle}{\emph{AAAI}}.
  \bibinfo{pages}{9466--9474}.
\newblock


\bibitem[\protect\citeauthoryear{Zhang, Xu, Wang, Wu, Keutzer, Gonzalez, and
  Tian}{Zhang et~al\mbox{.}}{2021}]%
        {noveld}
\bibfield{author}{\bibinfo{person}{Tianjun Zhang}, \bibinfo{person}{Huazhe Xu},
  \bibinfo{person}{Xiaolong Wang}, \bibinfo{person}{Yi Wu},
  \bibinfo{person}{Kurt Keutzer}, \bibinfo{person}{Joseph~E Gonzalez}, {and}
  \bibinfo{person}{Yuandong Tian}.} \bibinfo{year}{2021}\natexlab{}.
\newblock \showarticletitle{NovelD: A Simple yet Effective Exploration
  Criterion}. In \bibinfo{booktitle}{\emph{NeurIPS}}.
\newblock


\bibitem[\protect\citeauthoryear{Zheng, Chen, Wang, He, Hu, Chen, Fan, Gao, and
  Zhang}{Zheng et~al\mbox{.}}{2021}]%
        {emc}
\bibfield{author}{\bibinfo{person}{Lulu Zheng}, \bibinfo{person}{Jiarui Chen},
  \bibinfo{person}{Jianhao Wang}, \bibinfo{person}{Jiamin He},
  \bibinfo{person}{Yujing Hu}, \bibinfo{person}{Yingfeng Chen},
  \bibinfo{person}{Changjie Fan}, \bibinfo{person}{Yang Gao}, {and}
  \bibinfo{person}{Chongjie Zhang}.} \bibinfo{year}{2021}\natexlab{}.
\newblock \showarticletitle{Episodic Multi-agent Reinforcement Learning with
  Curiosity-driven Exploration}. In \bibinfo{booktitle}{\emph{NeurIPS}}.
\newblock


\bibitem[\protect\citeauthoryear{Zhou, Luo, Villella, Yang, Rusu, Miao, Zhang,
  Alban, Fadakar, Chen, et~al\mbox{.}}{Zhou et~al\mbox{.}}{2020}]%
        {auto_driver}
\bibfield{author}{\bibinfo{person}{Ming Zhou}, \bibinfo{person}{Jun Luo},
  \bibinfo{person}{Julian Villella}, \bibinfo{person}{Yaodong Yang},
  \bibinfo{person}{David Rusu}, \bibinfo{person}{Jiayu Miao},
  \bibinfo{person}{Weinan Zhang}, \bibinfo{person}{Montgomery Alban},
  \bibinfo{person}{Iman Fadakar}, \bibinfo{person}{Zheng Chen},
  {et~al\mbox{.}}} \bibinfo{year}{2020}\natexlab{}.
\newblock \showarticletitle{Smarts: Scalable multi-agent reinforcement learning
  training school for autonomous driving}.
\newblock \bibinfo{journal}{\emph{CoRR}}  \bibinfo{volume}{abs/2010.09776}
  (\bibinfo{year}{2020}).
\newblock


\end{thebibliography}


\clearpage
\appendix

\section{Details about Benchmarks}
\begin{figure}[ht]
    \centering
    \begin{minipage}{\linewidth}
        \begin{minipage}[c]{0.56\linewidth}
            \centering 
            \subfigure[c][SMAC environment]{
             \includegraphics[width=0.95\textwidth]{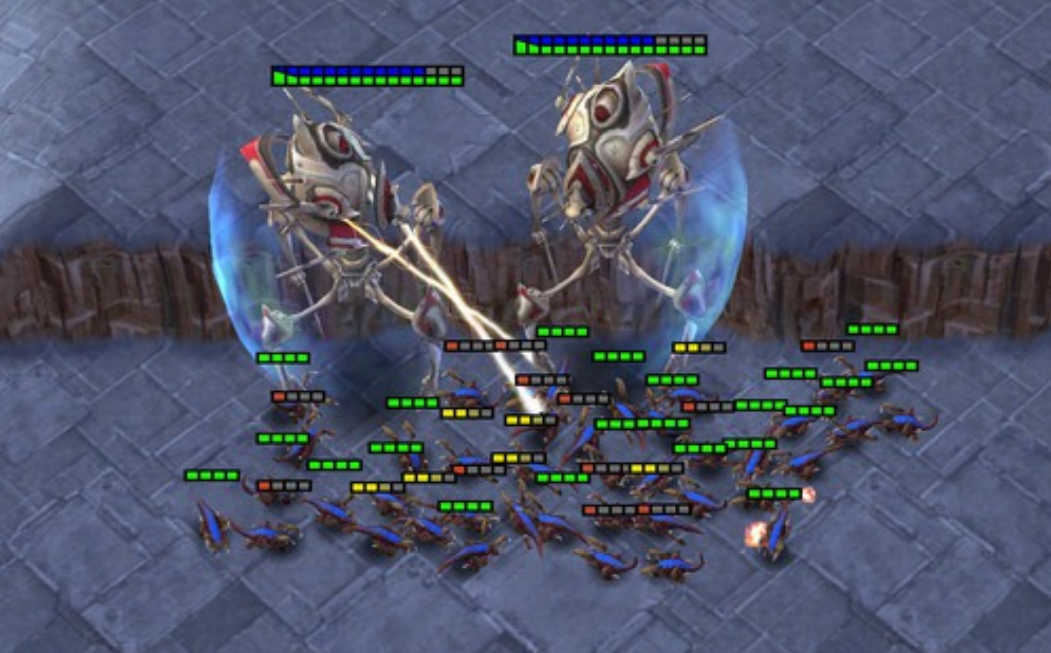}
            \label{fig:benchmark_smac}}
        \end{minipage} 
        \begin{minipage}[c]{0.41\linewidth}
            \centering
            \subfigure[c][LBF environment]{
            \includegraphics[width=0.95\textwidth]{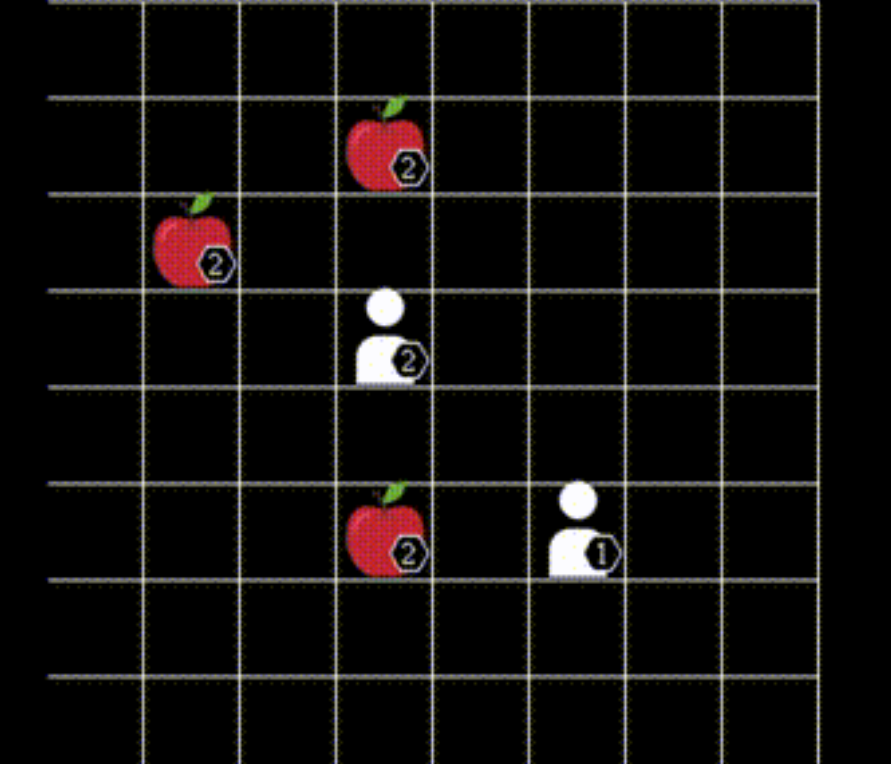}
            \label{fig:benchmark_lbf}}
        \end{minipage}
        \caption{Two benchmarks used in this paper.}
        \label{fig:appendix_benchmarks}
    \end{minipage}
\end{figure}

We use two testing environments in our paper~(Figure~\ref{fig:appendix_benchmarks}), including StarCraft II micromanagement benchmark~(SMAC)~\cite{smac} and Level Based Foraging~(LBF)~\cite{lbf}.

\textbf{StarCraft II micromanagement benchmark~(SMAC)} is an environment based on the famous game StarCraft II. In SMAC, there are some challenging combat scenarios, where a group of units are controlled by the Reinforcement Learning agents to battle an enemy army controlled by the game’s built-in scripted AI.

\textbf{Level Based Foraging~(LBF)} is another common MARL experiment environment. In LBF, a group of agents need to cooperate to eat the food. Each agent and each food has a level. Only when the sum level of the agents wanting to eat the food is not less than that of the food can they successfully eat it.  

\section{SMMAE Implementation Details}
\label{sec:imple_detail}

\begin{table*}[ht]
  \caption{The hyper-parameters of SMMAE in SMAC}
  \label{tab:hyperparameters}
  \centering
  \begin{tabular}{lc}
    \toprule
    Hyper-parameter & Value \\
    \midrule
    $\varepsilon$ anneal time & 50000  \\
    $\valpha$ increasing step scale $\lambda_\alpha$ & 10  \\
    Environment reward scale $\lambda_{exp}$ & 0.5 \\
    $\valpha$ update frequency $E_{\alpha}$ & 5 \\
    Exploration probability upper bound value $\alpha_{high}$     & 0.3 or 0.4 \\
    Exploration probability lower bound value $\alpha_{low}$     & 0.04 \\
    Exploration probability changing threshold $\gL_{threshold}$     & 0.6, 1.2, 1.4 or 1.5      \\
    Observation VAE model encoder/decoder hidden layer size     & 64    \\
    Observation VAE model latent dimension & 32 \\
    Observation VAE model learning rate & 0.01 \\
    \bottomrule
  \end{tabular}
\end{table*}

The implementation of the baselines is based on PyMARL~\cite{smac}. For QPLEX~\cite{qplex}, RODE~\cite{rode}, and EMC~\cite{emc}, we use the source code with the same hyper-parameters as they use in SMAC~\cite{smac}. For VDN~\cite{vdn} and QMIX~\cite{qmix}, we use the same hyper-parameters as that in SMAC~\cite{smac} and replace RMSprop optimizer with Adam optimizer using default hyper-parameters. In our experiments, SMMAE is based on QMIX with Adam optimizer. The attention key size and the attention value size are all 32. The other hyper-parameters are shown in Table~\ref{tab:hyperparameters}. Each $\alpha_i$ starts at $1.0$ and decays linearly to $0.05$ at $\varepsilon$ anneal time as that in common implementation. After the anneal time, each $\alpha_i$ will be updated according to Eq.~3
every $E_\alpha$ episodes. The tasks are trained on NVIDIA RTX 3090 GPU, and each task needs about 10 to 22 hours.

\begin{figure}[ht]
    \centering
    \includegraphics[width=0.7\linewidth]{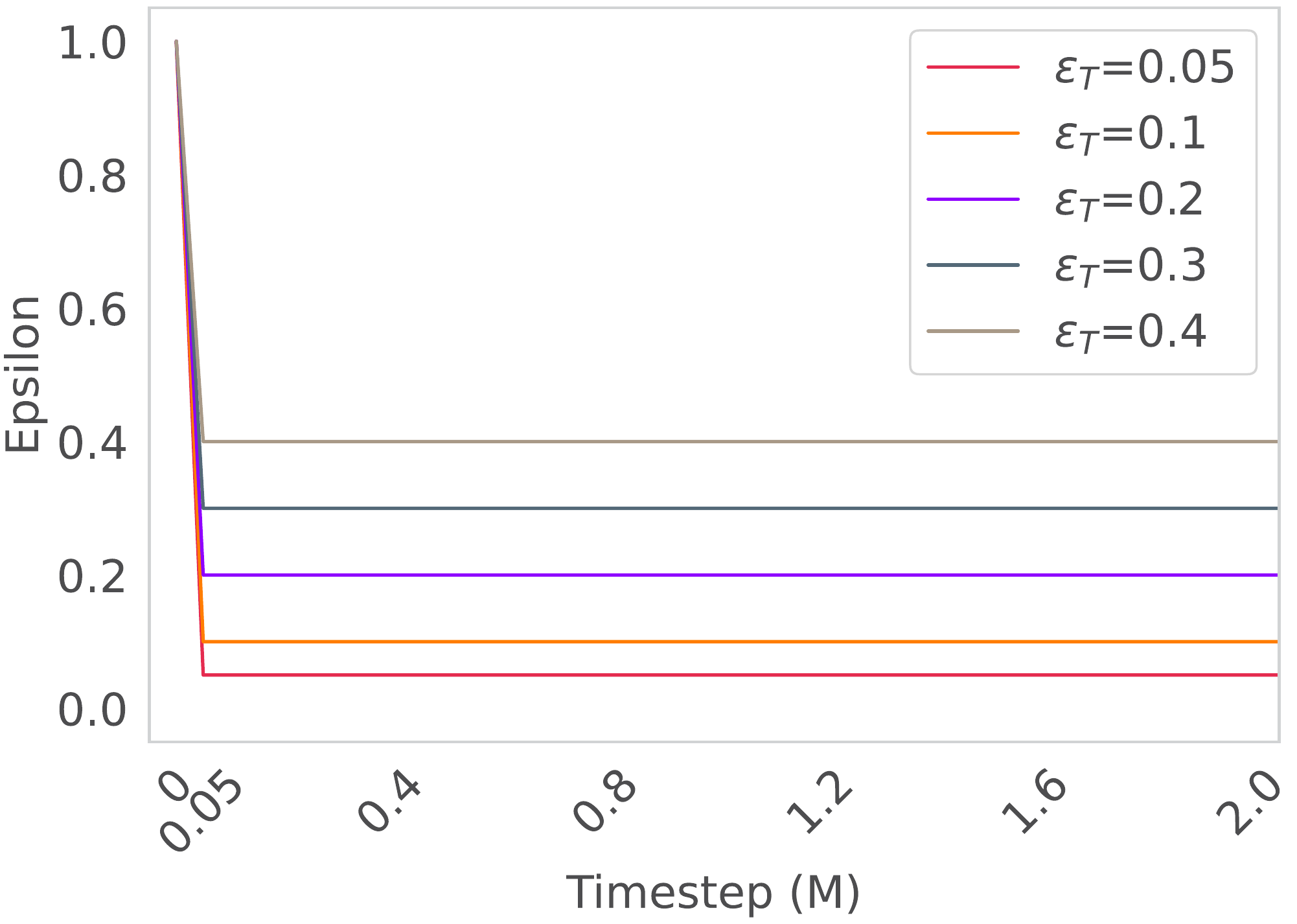}
    \caption{Effects of different hyper-parameter $\varepsilon_T$ on $\varepsilon$ value.}
 
    \label{fig:epsilon_finish_value_show}
\end{figure}

\section{Ablation Study of Epsilon}
\label{sec:epsilon_finish_ablation}

\begin{figure}[ht]
\centering
		\centerline{\includegraphics[width=1.0\linewidth]{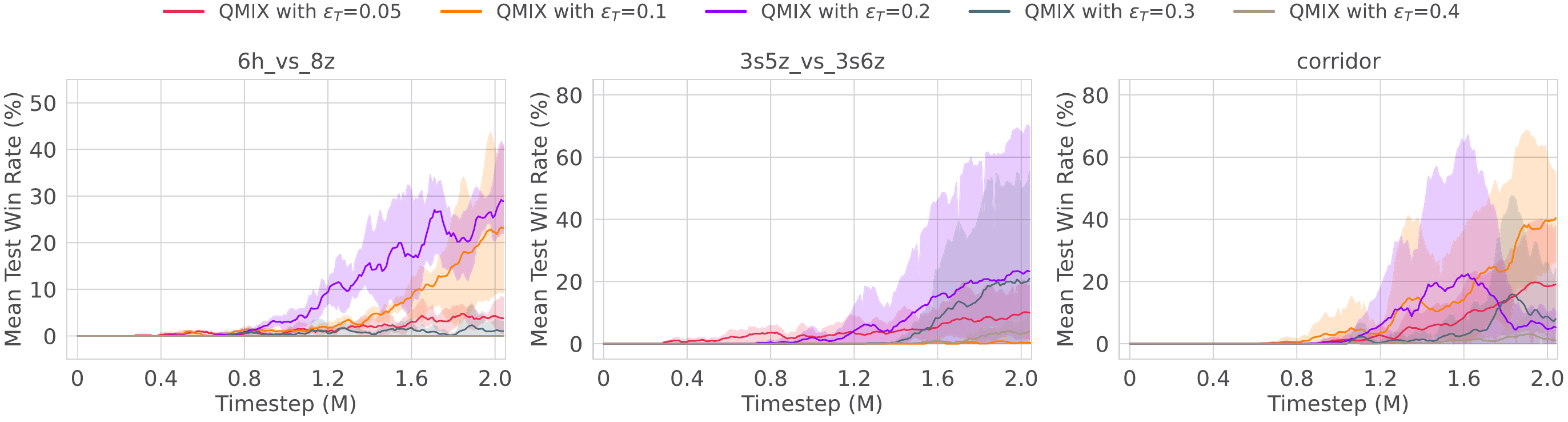}}
		\caption{Results of QMIX with different $\varepsilon_T$ on three super hard maps in the SMAC environments.}
		\label{fig:qmix_epsilon_finish_ablation}
\end{figure}

Figure~\ref{fig:qmix_epsilon_finish_ablation} shows the performance of QMIX with different epsilon finish values on 3 super hard maps in SMAC. These 3 maps are so hard that only a few algorithms can obtain a high test win rate in 2 million timesteps. The original QMIX~\cite{qmix} adopts RMSprop optimizer, and the epsilon finish value after $\varepsilon$ anneal time is $0.05$. In our experiments, we adopt Adam optimizer and change the epsilon finish value after anneal time. It shows that QMIX can achieve very different performances by only changing the hyper-parameter epsilon finish value $\varepsilon_T$. QMIX with $\varepsilon_T=0.2$ has better performance on \emph{6h\_vs\_8z} and \emph{3s5z\_vs\_3s6z}, while QMIX with $\varepsilon_T=0.1$ has better performance on \emph{corridor}. It shows that QMIX is very sensitive to the hyper-parameter epsilon finish value.

\begin{figure}[ht]
\centering
		\centerline{\includegraphics[width=1.0\linewidth]{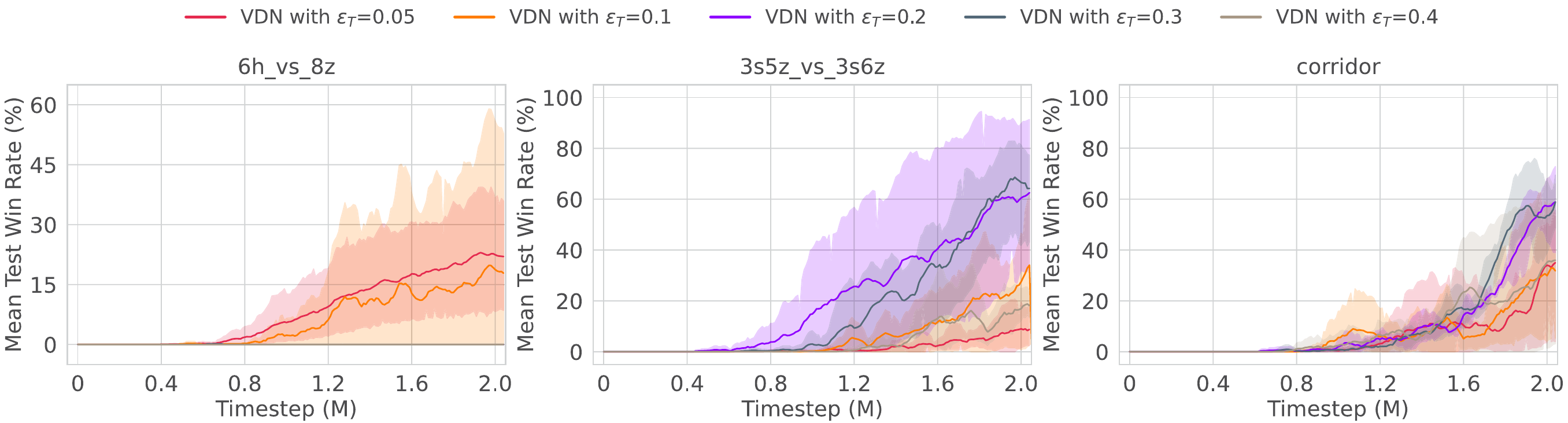}}
		\caption{Results of VDN with different $\varepsilon_T$ on three super hard maps in the SMAC environments.}
		\label{fig:vdn_epsilon_finish_ablation}
\end{figure}

\begin{figure}[ht]
    \centering
    \begin{minipage}{\linewidth}
        \begin{minipage}[c]{0.48\linewidth}
            \centering 
            \subfigure{
             \includegraphics[width=0.95\textwidth]{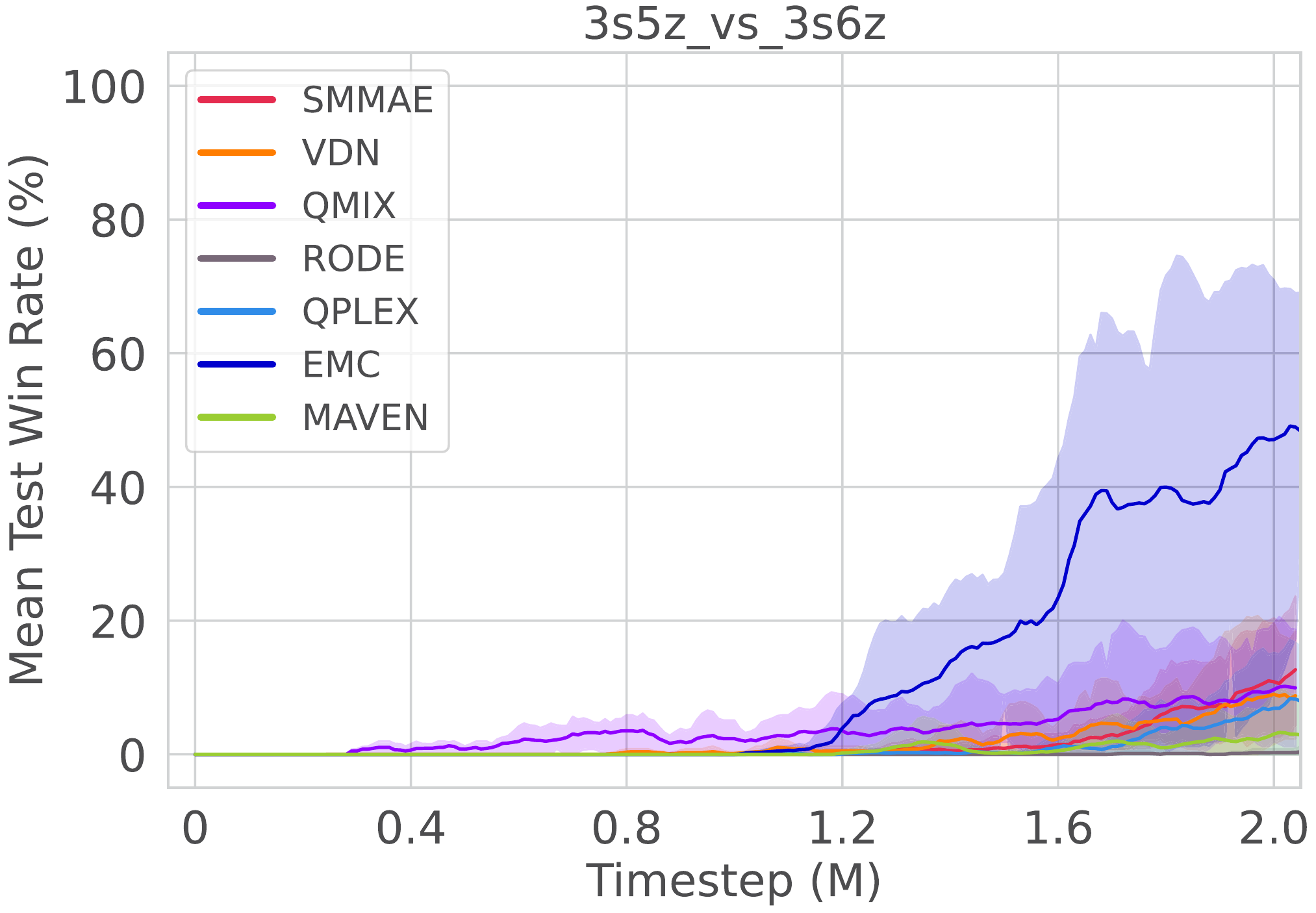}
            \label{fig:smac_sota_on_3s5z_vs_3s6z}}
        \end{minipage} 
        \begin{minipage}[c]{0.48\linewidth}
            \subfigure{
            \includegraphics[width=0.95\textwidth]{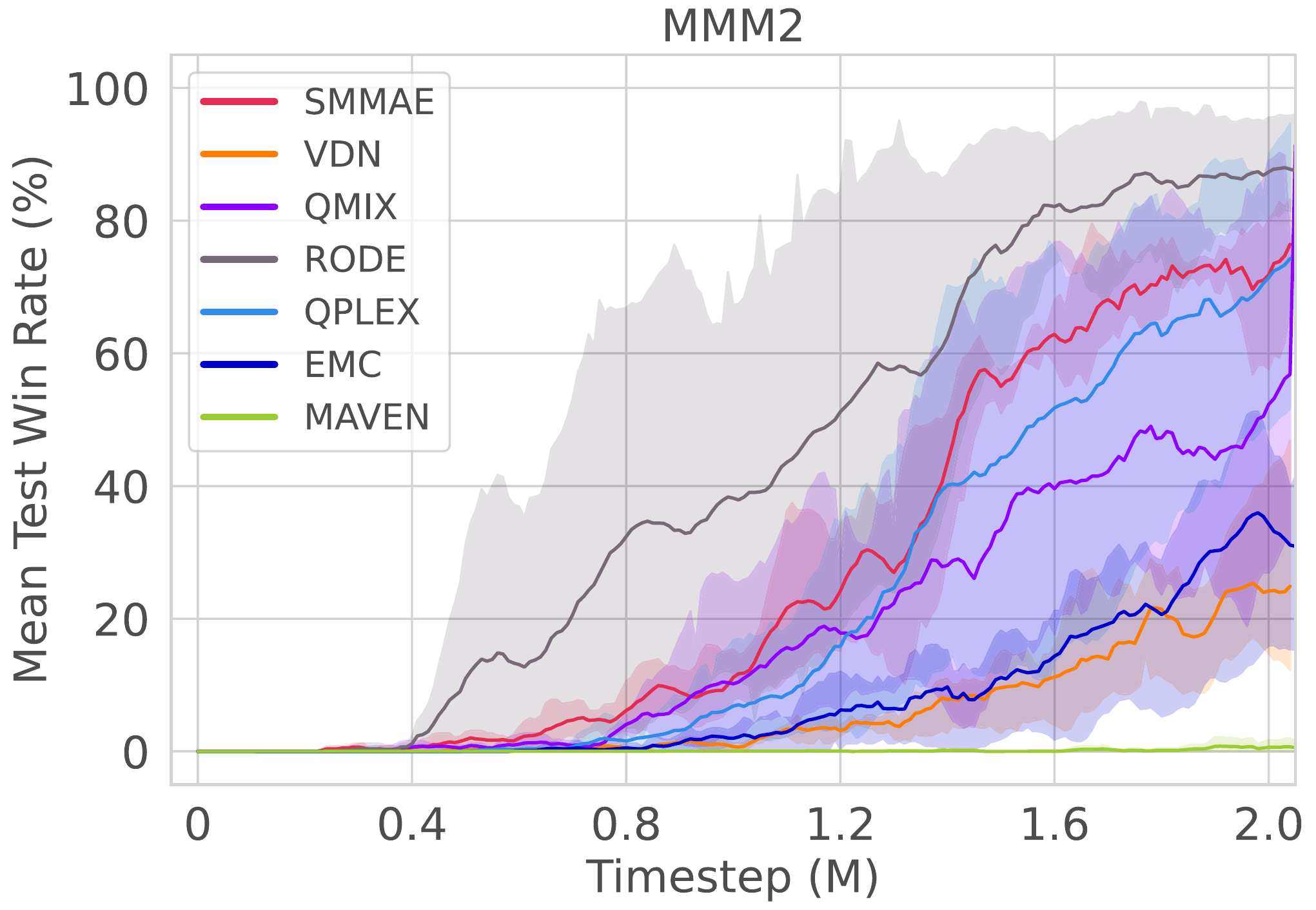}
            \label{fig:smac_sota_on_MMM2}}
        \end{minipage}
        \caption{Results on two super hard maps of SMAC.}
        \label{fig:appendix_smac_sota1}
    \end{minipage}
\end{figure}

VDN has similar results like that on QMIX. Figure~\ref{fig:vdn_epsilon_finish_ablation} shows the performance of VDN with different epsilon finish values on 3 super hard maps in SMAC. We also adopt Adam optimizer and change the epsilon finish value $\varepsilon_T$ here. VDN with $\varepsilon_T=0.05$ performs best in \emph{6h\_vs\_8z} and the mean test win rate is about $20\%$. While VDN with $\varepsilon_T=0.05$ performs best in \emph{3s5z\_vs\_3s6z} and the mean test win rate is about $60\%$. VDN with $\varepsilon_T=0.2$ and VDN with $\varepsilon_T=0.3$ have better performance in the hard map \emph{corridor}.

\section{Other Super Hard Scenarios in SMAC}
We also conduct experiments on the other 2 super hard maps in SMAC~(Figure~\ref{fig:appendix_smac_sota1}). In \emph{MMM2}, SMMAE can also obtain competitive results. In \emph{3s5z\_vs\_3s6z}, SMMAE can obtain better results than QMIX and some other baselines, but the final test win rate is not high. We assume that the reason is uniform distribution causes unexpected exploration.

\end{document}